\documentclass{article}

\usepackage{caption}

\usepackage{microtype}
\usepackage{graphicx}
\usepackage{subfigure}
\usepackage{booktabs} 

\usepackage{hyperref}

\usepackage[accepted]{icml2024}

\usepackage{amsmath}
\usepackage{amssymb}
\usepackage{mathtools}
\usepackage{amsthm}

\usepackage[export]{adjustbox} 
\usepackage{enumitem}

\usepackage{xcolor}         
\usepackage{multirow}

\usepackage{xspace}

\usepackage{graphicx}
\usepackage{colortbl}
\usepackage{wrapfig}
\usepackage{duckuments}
\usepackage{xcolor,pifont}

\usepackage{xcolor}
\usepackage[linesnumbered,ruled,vlined,algo2e]{algorithm2e}

\hypersetup{citecolor=MidnightBlue, linkcolor=BrickRed}

\definecolor{pinegreen}{rgb}{0.0, 0.47, 0.44}
\definecolor{cornellred}{rgb}{0.7, 0.11, 0.11}
\definecolor{cadmiumgreen}{rgb}{0.0, 0.42, 0.24}
\definecolor{spirodiscoball}{rgb}{0.06, 0.75, 0.99}
\definecolor{Red7}{rgb}{0.941, 0.243, 0.243}
\definecolor{Eqpink}{RGB}{241,241,214}
\definecolor{Green7}{RGB}{55, 178, 77}
\definecolor{aliceblue}{rgb}{0.91, 0.94, 0.97}
\definecolor{darkblue}{rgb}{0.83, 0.89, 0.97}
\definecolor{SJViolet}{RGB}{105,100,171}
\definecolor{SJRed}{RGB}{237,109,107}
\definecolor{tablegreen}{rgb}{0.91, 0.94, 0.97}

\hypersetup{
  linkcolor = cornellred,
  citecolor  = cadmiumgreen,
  colorlinks = true,
}

\newcommand{\algname}{\textbf{H}ierarchical textual \textbf{I}nversion for \textbf{Mol}ecular generation\xspace}
\newcommand{\Algname}{HI-Mol\xspace}
\newcommand{\ALgname}{Hierarchical textual Inversion for Molecular generation \xspace}
\newcommand{\ALGname}{HI-Mol\xspace}

\newcommand{\cmark}{\ding{51}}%
\newcommand{\xmark}{\ding{55}}%
\usepackage{tablefootnote}

\newcommand\sjline[1]{\textcolor{magenta}{#1}}

\newcommand{\stdv}[1]{\scriptsize$\pm$#1}

\usepackage[capitalize,noabbrev]{cleveref}

\theoremstyle{plain}

\theoremstyle{definition}

\theoremstyle{remark}

\usepackage[textsize=tiny]{todonotes}

\icmltitlerunning{Data-Efficient Molecular Generation with Hierarchical Textual Inversion}

\begin{document}

\twocolumn[
\icmltitle{Data-Efficient Molecular Generation with Hierarchical Textual Inversion}

\icmlsetsymbol{equal}{*}

\begin{icmlauthorlist}
\icmlauthor{Seojin Kim}{kaist}
\icmlauthor{Jaehyun Nam}{kaist}
\icmlauthor{Sihyun Yu}{kaist}
\icmlauthor{Younghoon Shin}{korea}
\icmlauthor{Jinwoo Shin}{kaist}

\end{icmlauthorlist}

\icmlaffiliation{kaist}{Korea Advanced Institute of Science and Technology (KAIST)}
\icmlaffiliation{korea}{Korea University}

\icmlcorrespondingauthor{Seojin Kim}{osikjs@kaist.ac.kr}

\icmlkeywords{Machine Learning, ICML}

\vskip 0.3in
]

\printAffiliationsAndNotice{}

\begin{abstract}
Developing an effective molecular generation framework even with a limited number of molecules is often important for its practical deployment, e.g., drug discovery, since acquiring task-related molecular data requires expensive and time-consuming experimental costs. 
To tackle this issue, we introduce \emph{\ALgname} (\ALGname), a novel data-efficient molecular generation method. \Algname is inspired by the importance of hierarchical information, e.g., both coarse- and fine-grained features, in understanding the molecule distribution. We propose to use multi-level embeddings to reflect such hierarchical features based on the adoption of the recent textual inversion technique in the visual domain, which achieves data-efficient image generation. Compared to the conventional textual inversion method in the image domain using a single-level token embedding, our multi-level token embeddings allow the model to effectively learn the underlying low-shot molecule distribution.
We then generate molecules based on the interpolation of the multi-level token embeddings.
Extensive experiments demonstrate the superiority of \ALGname with notable data-efficiency. 
For instance, on QM9, \ALGname outperforms the prior state-of-the-art method 
with 50$\times$ less training data. 
We also show the effectiveness of molecules generated by \ALGname in low-shot molecular property prediction. Code is available at \url{https://github.com/Seojin-Kim/HI-Mol}.
\end{abstract}

\section{Introduction}
\label{sec:intro}
Finding novel molecules has been a fundamental yet crucial problem in chemistry \citep{xue2019advances,xu2019deep} due to its strong relationship in achieving important applications, such as drug discovery \citep{segler2018generating, bongini2021molecular} and material design \citep{hamdia2019novel, tagade2019attribute}. 
However, generating molecules poses a challenge due to their highly complicated nature and the vast size of the input space \citep{drew2012size}.
To tackle this issue, several works have considered training deep generative models to learn the 
molecule distribution using large molecular datasets \citep{jin2018junction,NEURIPS2022_1160792e,ahn2022spanning,geng2023novo}. This is inspired by the recent breakthroughs of generative models in other domains, e.g., images and videos \citep{rombach2022high,singer2022make,yu2023video}, in learning high-dimensional data distributions. Intriguingly, such deep molecular generation methods have demonstrated reasonable 
performance \citep{jin2020hierarchical, ahn2022spanning,NEURIPS2022_1160792e} on the large-scale benchmarks \citep{ramakrishnan2014quantum,polykovskiy2020molecular} in finding chemically valid and novel molecules, showing great potential to solve the challenge.

\begin{figure*}[t]
\centering\small
\includegraphics[width=\linewidth]{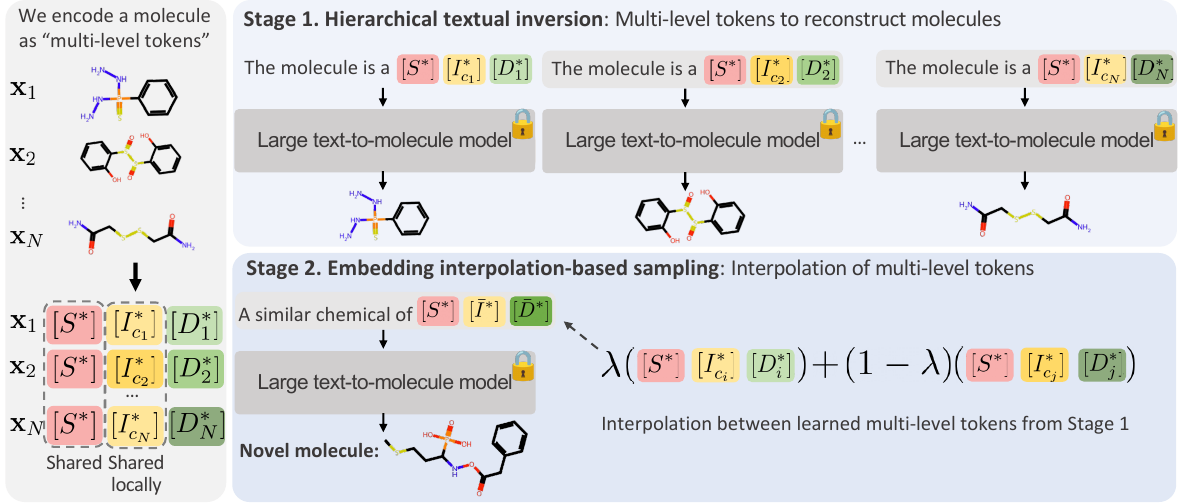}
\vspace{-0.25in}
\caption{
Overview of \Algname framework. (1) Hierarchical textual inversion: we encode low-shot molecules into multi-level token embeddings. (2) Embedding interpolation-based sampling: we generate novel molecules using interpolation of low-level token embeddings.
}
\label{fig:concept}
\vspace{-0.07in}
\end{figure*}

Unfortunately, existing molecular generation frameworks tend to fail in limited data regimes~\citep{guo2022data}. This restricts the deployment of existing approaches to practical scenarios, because task-related molecular data for the target real-world applications are often insufficient to train such molecular generative models. For example, drug-like molecules for a specific organ are inherently scarce in nature \citep{schneider2005computer,altae2017low}, and the drug-likeness of each candidate molecule should be verified through years of extensive wet experiments and clinical trials \citep{drews2000drug,hughes2011principles}. This time-consuming and labor-intensive data acquisition process of new task-related molecules limits the number of training data available for a model to learn the desired molecule distribution \citep{stanley2021fs}.
Thus, it is often crucial to develop an effective \emph{data-efficient molecular generation} framework, yet this direction has been overlooked in the field of deep molecular generation \citep{guo2022data}.

In this paper, we aim to address the aforementioned shortcomings of existing molecular generation frameworks in the low-shot regimes by designing a method to leverage knowledge in a limited number of molecules extensively. To this end, inspired by the chemical prior that molecules can be hierarchically clustered \citep{alexander2011bcl}, we introduce multi-level embeddings that capture coarse- and fine-grained features among low-shot molecules, where a selective assignment of the multi-level tokens for each molecule allows to incorporate hierarchical features of molecules.

We learn this hierarchical embedding in the token space of the recent text-to-molecule model \citep{edwards2022translation}. Specifically, we adopt \emph{textual inversion} \citep{gal2022image}---which learns low-shot image distribution by introducing a new single token within a text-to-image model. A key difference here is that the consideration of ``multi-level'' tokens (see Figure~\ref{fig:concept}) are essential in learning the low-shot molecule distribution due to the complicatedly structured nature of molecules compared to images (see Table~\ref{tab:motivation}).

\textbf{Contribution.} 
We introduce a novel data-efficient molecular generation method, coined \algname (\textbf{\Algname}). Specifically, \Algname is composed of the following components:
\begin{itemize}[topsep=1.0pt,itemsep=1.0pt,leftmargin=5.5mm]
    \item [$\bullet$] \textbf{Hierarchical textual inversion:} 
    We propose a molecule-specialized textual inversion scheme to capture the hierarchical information of molecules \citep{alexander2011bcl}. 
    In contrast to textual inversion for the visual domain that optimizes a single shared token on given training data, we design multi-level tokens for the inversion. 
    Thus, the shared token learns the common features among molecules and the low-level tokens learn cluster-specific or molecule-specific features.
    \vspace{-0.05in}
    \item [$\bullet$] \textbf{Embedding interpolation-based sampling:} We propose to use low-level tokens in addition to the shared token for molecular generation. In particular, we consider using the interpolation of low-level token embeddings. The mixing approach is designed to extensively utilize the information of given molecules, and thus effectively alleviates the issue of the limited number of molecules.
\end{itemize}

We extensively evaluate \Algname by designing several data-efficient molecular generation tasks on the datasets in the MoleculeNet benchmark \citep{wu2018moleculenet} and on the QM9 dataset \citep{ramakrishnan2014quantum}. 
For instance, in the HIV dataset in MoleculeNet, \Algname improves Frechet ChemNet Distance (FCD, lower is better; \citealp{preuer2018frechet}) as 20.2 $\to$ 16.6 from prior arts. 
On the QM9 dataset, \Algname already outperforms the previous state-of-the-arts, e.g., STGG \citep{ahn2022spanning} by 0.585 $\to$ 0.434 in FCD, 
with 50$\times$ less training data. {Finally, we validate the effectiveness of the molecules generated by our \Algname framework on the low-shot property prediction tasks in MoleculeNet.}

\section{Related Work}
\textbf{Molecular generation.}
Most molecular generation methods fall into three categories. 
First, graph-based methods \citep{jo2022score,hoogeboom2022equivariant,luo2022fast,zhang2023equivariant,vignac2023digress} formalize molecular generation as a graph generation problem by representing each molecule as an attributed graph. Next, fragment-based methods \citep{jin2018junction,NEURIPS2022_1160792e,geng2023novo} define a dictionary of chemically meaningful 
fragments, e.g., functional groups. A molecule is represented as a tree structure of dictionary elements and the distribution of connected fragments is then modeled. Finally, string-based methods \citep{gomez2016design,flam2022language,ahn2022spanning} utilize the Simplified Molecular-Input Line-Entry System \citep[SMILES,][]{weininger1988smiles} to write molecules as strings and learn the distribution of molecules in this string space. {Our method takes the string-based approach based on 
the recent large-scale text-to-molecule models that use the SMILES representation.}

\textbf{Hierarchical generation methods.} Recent molecular generation methods introduce the notion of hierarchy in molecular generation \citep{jin2020hierarchical,zhu2023molhf}. Specifically, they incorporate the hierarchy \emph{within} a single molecule, e.g., atom- and motif-level. However, they do not consider the hierarchy \emph{among} molecules, e.g., dataset-, cluster-, and molecule-level, which is indeed crucial to understand the molecular dataset, i.e., target distribution \citep{alexander2011bcl}. To overcome this limitation, we carefully design our multi-level embeddings to reflect the hierarchy \emph{among} the target molecules through our hierarchical tokens.

{\textbf{Molecular language model.}
{Following the recent progress in text-conditional generative models, e.g., text-to-text \citep{raffel2020exploring, touvron2023llama} and text-to-image \citep{ramesh2021zero,rombach2022high}, there exist several attempts to train text-to-molecule models, i.e., molecular language models~\citep{bagal2021molgpt,christofidellis2023unifying,liu2023multi}. Specifically, these works exploit popular language model architectures 
to have pre-trained models for molecules, based on the SMILES \citep{weininger1988smiles} representation 
that interprets a given molecule 
as a string. For instance, MolT5 \citep{edwards2022translation} proposes to fine-tune a large text-to-text language model, T5~\citep{raffel2020exploring}, with SMILES representations of large-scale molecular data and description-SMILES pair data to have a text-to-molecule model. Notably, it results in a highly effective pre-trained model for molecules, demonstrating superior performance across several text-to-molecule generation tasks. Building on its success, we mainly utilize the Large-Caption2Smiles model trained with this MolT5 approach for our goal of data-efficient molecular generation.\footnote{{We provide the results utilizing other text-to-molecule models \citep{christofidellis2023unifying,pei2023biot5} in Appendix~\ref{sup:model_ablation}.}}}

\textbf{Low-shot generation.}
In the field of generative models, there have been considerable efforts 
to design a low-shot generation framework for generating new samples from a given small number of data \citep{wang2018transferring,noguchi2019image}. 
Intriguingly, recent works on large-scale text-to-image diffusion models 
have surprisingly resolved this challenge, even enabling ``personalization'' of the model to a few in-the-wild images through 
simple optimization schemes that update only a few parameters of a pre-trained model~\citep{gal2022image,cohen2022my,wei2023elite}. In particular, textual inversion~\citep{gal2022image} exhibits that the personalization of large-scale text-to-image diffusion models with a small number of images can be achieved even with 
a very simple optimization of a single additional text token  
without updating any pre-trained model parameters.

{In contrast to the recent advances of low-shot generation in the image domain, developing a low-shot (or data-efficient) molecular generation framework is relatively under-explored despite its importance in practical applications \citep{altae2017low,guo2022data}. 
Hence, our method tackles this problem by designing a molecule-specific inversion method using the recent large-scale text-to-molecule models. Specifically, we incorporate the concept of ``hierarchy'' of molecular structures \citep{alexander2011bcl} into the textual inversion framework. Due to this unique motivation, our method effectively learns the molecule distribution with the shared concept in low-shot molecules with diverse molecular structures, while the applications of prior works, e.g., \citet{guo2022data}, are limited to structurally similar low-shot molecules such as monomers and chain-extenders (see Table~\ref{tab:moleculenet_few} for comparison).}

\section{\Algname: Hierarchical Textual Inversion for Molecular Generation}

In Section~\ref{subsec:overview}, we provide an overview of our problem and the main idea. In Section~\ref{subsec:prelim}, we provide descriptions of textual inversion to explain our method. In Section~\ref{subsec:molecular_inversion}, we provide a component-wise description of our method.

\subsection{Problem Description and Overview}
\label{subsec:overview}

{
We formulate our problem of \emph{data-efficient molecular generation} as follows. Consider a given molecular data $\mathcal{M} \coloneqq \{ \mathbf{x}_n \}_{n=1}^{N}$, where each molecule $\mathbf{x}_n$ is drawn from an unknown task-related molecule distribution $p(\mathbf{x} | \mathbf{c})$. 
{Here, $\mathbf{c}$ represents the 
common underlying chemical concept among molecules in the dataset for the target task, e.g., blood-brain barrier permeability.
} We aim to learn a model distribution $p_{\mathrm{model}}(\mathbf{x})$ that matches $p(\mathbf{x} | \mathbf{c})$, where the number of molecules $N$ 
is small, e.g., $N=691$ in the BACE dataset.}

{To solve this problem, we take the recent approach of textual inversion~\citep{gal2022image} from the text-to-image model literature---a simple yet powerful technique in low-shot image generation that learns a common concept in given images as a single token in the text embedding space. Motivated by its success, we aim to learn the common chemical concept of molecules as text tokens and use them for our goal of data-efficient generation. However, 
we find that the na\"ive applications of inversion fail in molecules (see Table~\ref{tab:motivation}). 
Unlike images, molecules with similar semantics often have entirely different structures (see Figure~\ref{fig:concept}), making it difficult to simply learn the common concept as a single text token.
Our contribution lies in resolving this challenge by adopting {molecule-specific priors, i.e., hierarchy,} into the framework to enjoy the power of textual inversion techniques in achieving data-efficient molecular generation.}

\subsection{Preliminary: Textual Inversion}
\label{subsec:prelim}

{Recent text-to-image generation methods have proposed textual inversion~\citep{gal2022image}, which aims to learn a common concept $\mathbf{c}$, i.e., the distribution $p(\mathbf{x}|\mathbf{c})$, from a small set of images and use it for the concept-embedded (or personalized) generation. To achieve this, they optimize a \emph{single} text embedding of a new token $[S^\ast]$ shared among images to learn $\mathbf{c}$ using a frozen pre-trained text-to-image model $f_{\tt t2i}$. Specifically, they put $[S^\ast]$ with a short text description, e.g., ``$\mathsf{A\,\, photo\,\,of\,\,}[S^\ast]$'', as the text prompt to $f_{\tt t2i}$, and then optimize this token embedding using given low-shot images with the exact same training objective that is used for training $f_{\tt t2i}$. 
{We propose to adapt the textual inversion method into the data-efficient molecular generation framework based on the recently proposed large-scale pre-trained text-to-molecule model \citep{edwards2022translation}.
}}

\subsection{Detailed Description of \Algname}
\label{subsec:molecular_inversion}
\begin{table}[t]
\caption{The ratio of valid generated molecules (Validity) based on na\"ive adoption of inversion methods in visual domain for data-efficient molecular generation in the HIV dataset \citep{wu2018moleculenet}. 
}
\vspace{0.05in}
\small
\label{tab:motivation}
\begin{center}
\begin{tabular}{c|c}
\toprule

            Inversion method & Validity (\%)  \\ \midrule

 Textual Inversion \citep{gal2022image} & 0.4 \\ 
 DreamBooth \citep{ruiz2022dreambooth} & 0.0 \\
 
\bottomrule

\end{tabular}
\end{center}
\end{table}
{\textbf{Failure of conventional inversion in molecules.} We conduct an experiment to explore the applicability of existing inversion methods \citep{gal2022image,ruiz2022dreambooth} in the visual domain for our goal of data-efficient molecular generation. These methods use a text prompt with a single shared token $[S^\ast]$ for the inversion of low-shot images based on the recent text-to-image models \citep{rombach2022high,saharia2022photorealistic}. Similarly, we apply their training objectives to low-shot molecules with a molecular language model \citep{edwards2022translation}. In contrast to the success in the low-shot image generation tasks, in Table~\ref{tab:motivation}, we show that such na\"ive applications of inversion methods fail in the molecular domain, i.e., they do not generate enough valid molecules, which motivates us to design a molecule-specialized inversion for data-efficient molecular generation.}

{\textbf{Hierarchical textual inversion.}}
{We first propose a molecule-specific textual inversion to learn the desired distribution of low-shot molecules. Unlike
the original textual inversion \citep{gal2022image} that assumes a single shared token $[S^\ast]$ only, we propose to use ``hierarchical'' tokens $[S^\ast], \{[I^\ast_k]\}_{k=1}^K, \{[D^\ast_n]\}_{n=1}^N$ (with parametrizations $\theta\coloneqq (\mathbf{s}, \{\mathbf{i}_k\}_{k=1}^K, \{\mathbf{d}_n\}_{n=1}^N)$) by introducing additional intermediate tokens $\{[I^\ast_k]\}_{k=1}^K$ and detail tokens $\{[D^\ast_n]\}_{n=1}^N$ (with $K<N$) to extensively incorporate the hierarchical features in training molecules. 
For instance, intermediate and detail tokens enable to learn different level of features, i.e., cluster-wise and molecule-wise, respectively.}

\begin{table*}[t]

\caption{Quantitative results of the generated molecules on the three datasets (HIV, BBBP, BACE) in the MoleculeNet benchmark \citep{wu2018moleculenet}. We mark in Grammar if the method explicitly exploits the grammar of molecular data and thus yields a high Valid. score. The Active. score is averaged over three independently pre-trained classifiers. We report the results using the 500 non-overlapping generated molecules to the training dataset. We set the highest score in bold. $\uparrow$ and $\downarrow$ indicate higher and lower values are better, respectively.} 
\vspace{-0.05in}
\label{tab:moleculenet}
\small
\begin{center}
\resizebox{1.0\textwidth}{!}{%

\begin{tabular}{clcccccccc}
\toprule

                        Dataset & Method  & Class & Grammar & Active. $\uparrow$ & FCD $\downarrow$ & NSPDK $\downarrow$ & Valid. $\uparrow$ & Unique. $\uparrow$ & Novelty $\uparrow$ \\ \midrule

\multirow{9.5}{*}{HIV}

& GDSS \citep{jo2022score}    & Graph           & \textcolor{SJRed}{\xmark}      &  \phantom{0}0.0         & 34.1 & 0.080 & 69.4  & \textbf{100}    & \textbf{100}     \\
& DiGress \citep{vignac2023digress}      &  Graph&\textcolor{SJRed}{\xmark}            & \phantom{0}0.0     & 26.2  & 0.067   & 17.8   & \textbf{100}   & \textbf{100} \\  
& JT-VAE \citep{jin2018junction}  & Fragment&\textcolor{SJViolet}{\cmark}  & \phantom{0}0.0 & 38.8 & 0.221 & \textbf{100} & 25.4 & \textbf{100} \\
& PS-VAE \citep{NEURIPS2022_1160792e}  & Fragment&\textcolor{SJViolet}{\cmark} & \phantom{0}{3.7} &  21.8 & 0.053 & \textbf{100} & 91.4 & \textbf{100}\\
&MiCaM \citep{geng2023novo} & Fragment & \textcolor{SJViolet}{\cmark} & \phantom{0}{3.4} &  20.4 & 0.037 & \textbf{100} & 81.6 & \textbf{100}\\
& CRNN \citep{segler2018generating}&  SMILES & \textcolor{SJRed}{\xmark}  & {\phantom{0}3.3} & 29.7 & 0.064 & 30.0 & \textbf{100} & \textbf{100} \\
& STGG \citep{ahn2022spanning} & SMILES& \textcolor{SJViolet}{\cmark}  & \phantom{0}1.6 & {20.2} & {0.033} & \textbf{100} & 95.8 & \textbf{100}\\
 \cmidrule(lr){2-10}
& \cellcolor{tablegreen}\textbf{\Algname (Ours)}& \cellcolor{tablegreen}SMILES & \cellcolor{tablegreen}\textcolor{SJRed}{\xmark}  & \cellcolor{tablegreen}\textbf{11.4} & \cellcolor{tablegreen}{19.0} & \cellcolor{tablegreen}\textbf{0.019} & \cellcolor{tablegreen}60.6 & \cellcolor{tablegreen}94.1 & \cellcolor{tablegreen}\textbf{100}\\ 
& \cellcolor{tablegreen}\textbf{\Algname (Ours)}& \cellcolor{tablegreen}SMILES & \cellcolor{tablegreen}\textcolor{SJViolet}{\cmark}  & \cellcolor{tablegreen}\textbf{11.4} & \cellcolor{tablegreen}\textbf{16.6} & \cellcolor{tablegreen}\textbf{0.019} & \cellcolor{tablegreen}\textbf{100} & \cellcolor{tablegreen}95.6 & \cellcolor{tablegreen}\textbf{100}\\

\midrule

\multirow{9.5}{*}{BBBP}

& GDSS \citep{jo2022score}     
& Graph& \textcolor{SJRed}{\xmark}  & \phantom{0}0.0 & 35.7 & 0.065 & 88.4  & {99.2}   & \textbf{100} \\
& DiGress \citep{vignac2023digress}   
 & Graph& \textcolor{SJRed}{\xmark}      & \phantom{0}8.2 & {17.4}    & 0.033     & 43.8     & 94.6      & \textbf{100}      \\ 

& JT-VAE \citep{jin2018junction} 
& Fragment& \textcolor{SJViolet}{\cmark}  & 80.6 & 37.4 & 0.202 & \textbf{100} & 10.8 & \textbf{100} \\
& PS-VAE \citep{NEURIPS2022_1160792e} 
& Fragment& \textcolor{SJViolet}{\cmark}  & 84.9& 17.3 & 0.039 & \textbf{100} & 91.6 & \textbf{100}\\
& MiCaM \citep{geng2023novo} 
& Fragment& \textcolor{SJViolet}{\cmark}  & 82.0& 14.3 & 0.021 & \textbf{100} & 89.4 & \textbf{100}\\

& CRNN \citep{segler2018generating} 
& SMILES& \textcolor{SJRed}{\xmark}  & 88.8 & 20.2 & 0.026 & 54.0 & \textbf{100} & \textbf{100} \\
& STGG \citep{ahn2022spanning} 
&  SMILES& \textcolor{SJViolet}{\cmark}  & {89.1} &{14.4} & {0.019} & 99.8 & 95.8 & \textbf{100}\\ 

\cmidrule(lr){2-10}
& \cellcolor{tablegreen}\textbf{\Algname (Ours)}  
& \cellcolor{tablegreen}SMILES& \cellcolor{tablegreen}\textcolor{SJRed}{\xmark}                 &  \cellcolor{tablegreen}{94.4}      & \cellcolor{tablegreen}{11.2} & \cellcolor{tablegreen}{0.011} & \cellcolor{tablegreen}78.8  & \cellcolor{tablegreen}92.9   & \cellcolor{tablegreen}\textbf{100}   \\ 
& \cellcolor{tablegreen}\textbf{\Algname (Ours)}  
& \cellcolor{tablegreen}SMILES& \cellcolor{tablegreen}\textcolor{SJViolet}{\cmark}                 &  \cellcolor{tablegreen}\textbf{94.6}      & \cellcolor{tablegreen}\textbf{10.7} & \cellcolor{tablegreen}\textbf{0.009} & \cellcolor{tablegreen}\textbf{100}  & \cellcolor{tablegreen}94.2   & \cellcolor{tablegreen}\textbf{100}   \\ 
\midrule
\multirow{9.5}{*}{BACE}

& GDSS \citep{jo2022score}&Graph  & \textcolor{SJRed}{\xmark}&  \phantom{0}9.1      & 66.0 & 0.205 & 73.4  & \textbf{100}    & \textbf{100}     \\

& DiGress \citep{vignac2023digress} & Graph &\textcolor{SJRed}{\xmark} &  21.1 & {26.7}  &  {0.102} & 16.4  & \textbf{100}  & \textbf{100}            \\

& JT-VAE \citep{jin2018junction} & Fragment &\textcolor{SJViolet}{\cmark} &  40.4 & 49.1 & 0.304 & \textbf{100} & 13.0 & \textbf{100}\\
& PS-VAE \citep{NEURIPS2022_1160792e} & Fragment &\textcolor{SJViolet}{\cmark} & 57.3 & 30.2 & 0.111& \textbf{100} & 75.6 & \textbf{100} \\

& MiCaM \citep{geng2023novo} & Fragment &\textcolor{SJViolet}{\cmark} & 56.2 & 18.5 & 0.060 & \textbf{100} & 64.2 & \textbf{100} \\
  
& CRNN \citep{segler2018generating} & SMILES & \textcolor{SJRed}{\xmark} & {79.0} &  21.7 & 0.066 & 38.0 & \textbf{100} & \textbf{100} \\
& STGG \citep{ahn2022spanning} &SMILES &\textcolor{SJViolet}{\cmark}&  42.9& {17.6} & {0.053} & \textbf{100} & 94.8 & \textbf{100}\\

\cmidrule(lr){2-10}
& \cellcolor{tablegreen}\textbf{\Algname (Ours)}  & \cellcolor{tablegreen}SMILES& \cellcolor{tablegreen}\textcolor{SJRed}{\xmark}                  &   \cellcolor{tablegreen}\textbf{81.0}  & \cellcolor{tablegreen}{16.4} & \cellcolor{tablegreen}{0.052}  &  \cellcolor{tablegreen}71.0  &  \cellcolor{tablegreen}69.9 &    \cellcolor{tablegreen}\textbf{100}   \\ 
& \cellcolor{tablegreen}\textbf{\Algname (Ours)}  & \cellcolor{tablegreen}SMILES& \cellcolor{tablegreen}\textcolor{SJViolet}{\cmark}                  &   \cellcolor{tablegreen}{80.4}  & \cellcolor{tablegreen}\textbf{14.0} & \cellcolor{tablegreen}\textbf{0.039}  &  \cellcolor{tablegreen}\textbf{100} &  \cellcolor{tablegreen}74.4 &    \cellcolor{tablegreen}\textbf{100}   \\ 
\bottomrule

\end{tabular}

}
\end{center}
\end{table*}

To learn these hierarchical tokens, we consider a frozen text-to-molecule model $f$, e.g., Large-Caption2Smiles \citep{edwards2022translation}, to apply our proposed hierarchical textual inversion objective. 
Specifically, we optimize $\theta$ by minimizing the following objective on the given molecular dataset $\mathcal{M}$:
\begin{align}
\label{eq:training}
    \nonumber
    \mathcal{L}(\theta;\mathbf{x}_n)\coloneqq \min_{k \in [1, K]}
    \mathcal{L}_{\mathtt{CE}} 
    \Big(
    f
    (\text{``}&{\mathsf{The\,\,molecule\,\,is\,\,a\,\,}}\\ 
    &{[S^\ast][I^\ast_{k}][D^\ast_n]\text{''}})
    ,\,\, 
    \mathbf{x}_n
    \Big),
\end{align}
{where $\mathcal{L}_{\mathtt{CE}}$ denotes cross-entropy loss and $\mathbf{x}_n$ is represented as its corresponding SMILES~\citep{weininger1988smiles} string.\footnote{Our method is also applicable to any future text-to-molecule models that represent $\mathbf{x}_n$ as graphs or 3D point clouds by replacing $\mathcal{L}_{\mathtt{CE}}$ with an appropriate objective to reconstruct $\mathbf{x}_n$.}  
Thus, after training, each molecule $\mathbf{x}_n$ is interpreted as text tokens $[S^\ast][I^\ast_{c_n}][D^\ast_n]$, where {we assign the intermediate token index $c_n \in [1,K]$ (for given $\mathbf{x}_n$ and the corresponding $[D^\ast_n]$) 
during optimization to minimize the training objective $\mathcal{L}$ (see Eq. (\ref{eq:training})).}
We note that the selection of $[I^\ast_{c_n}]$ is achieved in an unsupervised manner so that it does not require specific information about each molecule. Intriguingly, we find that $[I^\ast_{c_n}]$ can learn some of the informative cluster-wise features through this simple selection scheme although we have not injected any prior chemical knowledge of the given molecular data (see Figure~\ref{fig:cluster} for an example).}

{Our ``multi-level'' token design is particularly important for the successful inversion with molecules because molecules have a different nature from images that are typically used in the existing textual inversion method. Image inputs in the conventional textual inversion are visually similar, e.g., pictures of the same dog with various poses, whereas molecules often have entirely different structures even if they share the common chemical concept, e.g., activeness on the blood-brain membrane permeability \citep{wu2018moleculenet}. This difference makes it difficult to learn the common concept as a simple single token; we mitigate it by adopting hierarchy in the inversion scheme by incorporating the principle of the chemistry literature highlighting that molecular data can be clustered hierarchically \citep{alexander2011bcl}.}

{\textbf{Embedding interpolation-based sampling.}}
We propose a sampling strategy from the learned distribution via our hierarchical textual inversion framework. We find that the na\"ive application of the sampling schemes used in existing textual inversion for images, e.g., putting a text prompt including the shared token $[S^\ast]$ such as ``$\mathsf{A\,\,similar\,\,chemical\,\,of\,\,} [S^\ast]$'' into the molecular language model $f$, 
does not show reasonable performance in molecular generation (see Table~\ref{tab:motivation}).

{To alleviate this issue, we propose to utilize the learned hierarchy information of molecules obtained in our textual inversion, i.e., intermediate tokens $\{[I^\ast_k]\}_{k=1}^K$ and detail tokens $\{[D^\ast_n]\}_{n=1}^N$, to sample from our target distribution. We consider the interpolation of each of $[I^\ast_{c_n}]$ and $[D^\ast_n]$ in the sampling process. Specifically, we sample a novel molecule with random molecule indices $i,j$ uniformly sampled from $[1,N]$ and a coefficient $\lambda$ drawn from a pre-defined prior distribution $p(\lambda)$ (see Appendix~\ref{appen:method} for our choice of $p(\lambda$)):}
\begin{align}
\label{eq:lambda}
\nonumber
\big(\bar{\mathbf{i}}, \bar{\mathbf{d}}\big) &\coloneqq \lambda \big(\mathbf{i}_{c_i}, \mathbf{d}_i\big) + (1-\lambda) \big(\mathbf{i}_{c_j}, \mathbf{d}_j\big), \\
\mathbf{x} &\coloneqq f \big(\text{``}\mathsf{A\,\,similar\,\,chemical\,\,of\,\,}[S^\ast][\bar{I}^\ast][\bar{D}^\ast]\text{''}\big),
\end{align}
where $[\bar{I}^\ast], [\bar{D}^\ast]$ indicate that we pass interpolated token embeddings $\bar{\mathbf{i}}, \bar{\mathbf{d}}$ to $f$, respectively, and $c_n \in [1, K]$ is an index of the intermediate token of a given molecule $\mathbf{x}_n$, i.e., an intermediate token index that minimizes the training objective in Eq.~(\ref{eq:training}).\footnote{We simply set the number of clusters $K$ to 10 in our experiments. Please see Appendix~\ref{sup:abla} for the analysis of $K$.} This additional consideration of low-level tokens $\{[I^\ast_k]\}_{k=1}^K, \{[D^\ast_n]\}_{n=1}^N$ (as well as $[S^\ast]$) encourages the sampling process to exploit the knowledge from the molecular dataset extensively, mitigating the issue of scarcity of molecules that lie in our desired molecule distribution 
and thus enables to generate high-quality molecules. We provide qualitative analysis of our sampling scheme in Appendix~\ref{appen:interpolation_analysis}.

\begin{table*}[t]
\caption{Qualitative results of the generated molecules on the two datasets (HIV, BBBP) of the MoleculeNet benchmark \citep{wu2018moleculenet}. 
We visualize the generated molecules from each method that has the maximum Tanimoto similarity with a given anchor molecule. 
We report the similarity below each visualization of the generated molecule. 
We set the highest similarity in bold.}
\vspace{-0.07in}
\label{tab:qual_result}
\begin{center}
\resizebox{1.0\textwidth}{!}{
\small
\begin{tabular}{c|cccc|c}
\toprule
Dataset               & DiGress~\citep{vignac2023digress} & 
MiCaM \citep{geng2023novo} &
 STGG~\citep{ahn2022spanning} & \textbf{\ALGname~(Ours)} & Train   
\\ \midrule
\multirow{2.5}{*}{HIV}   &  
\includegraphics[height=0.66in,valign=c]{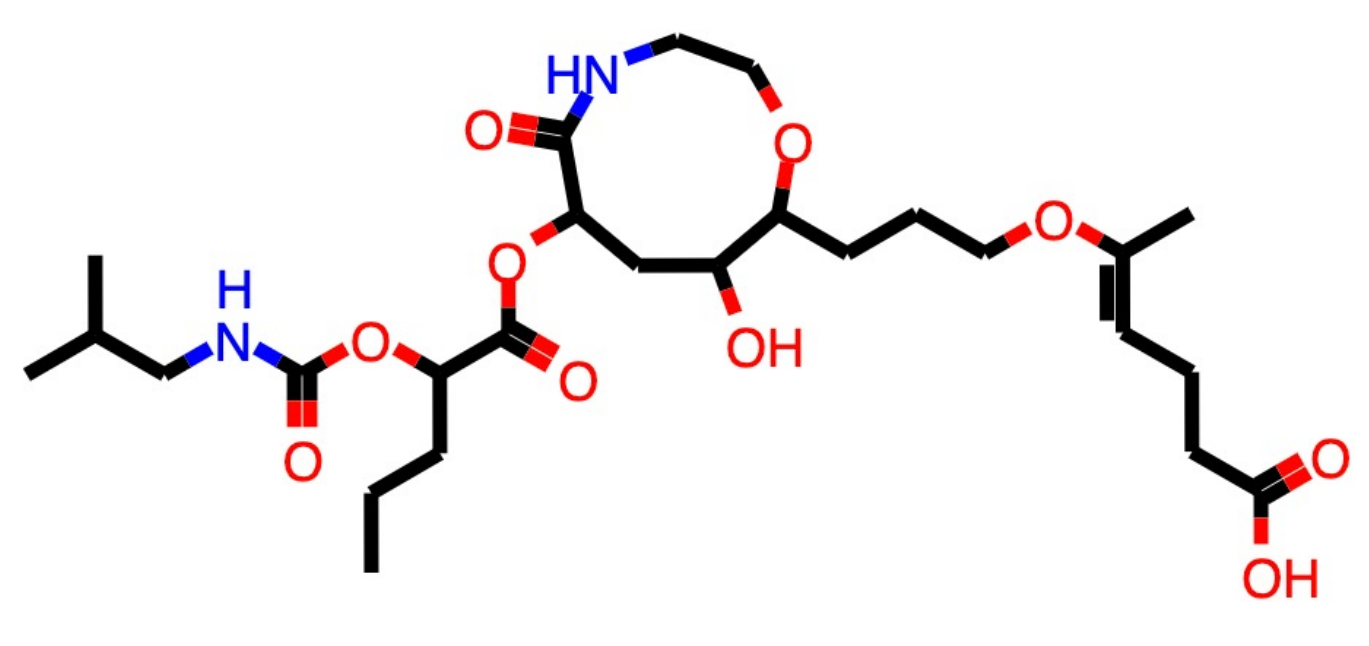} & 
\includegraphics[height=0.66in,valign=c]{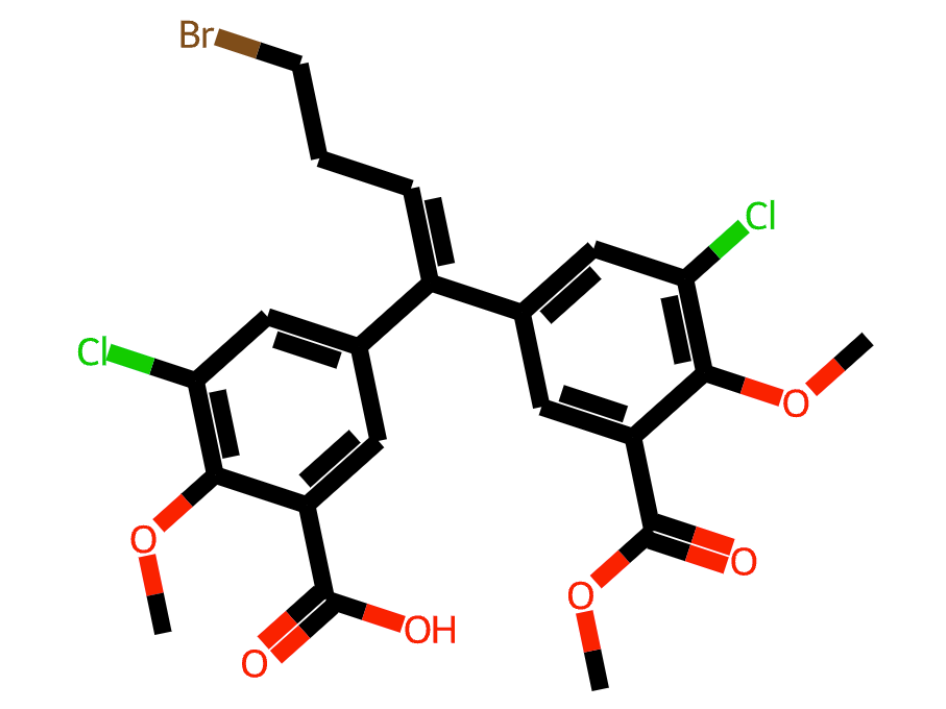}   & 
\includegraphics[height=0.66in,valign=c]{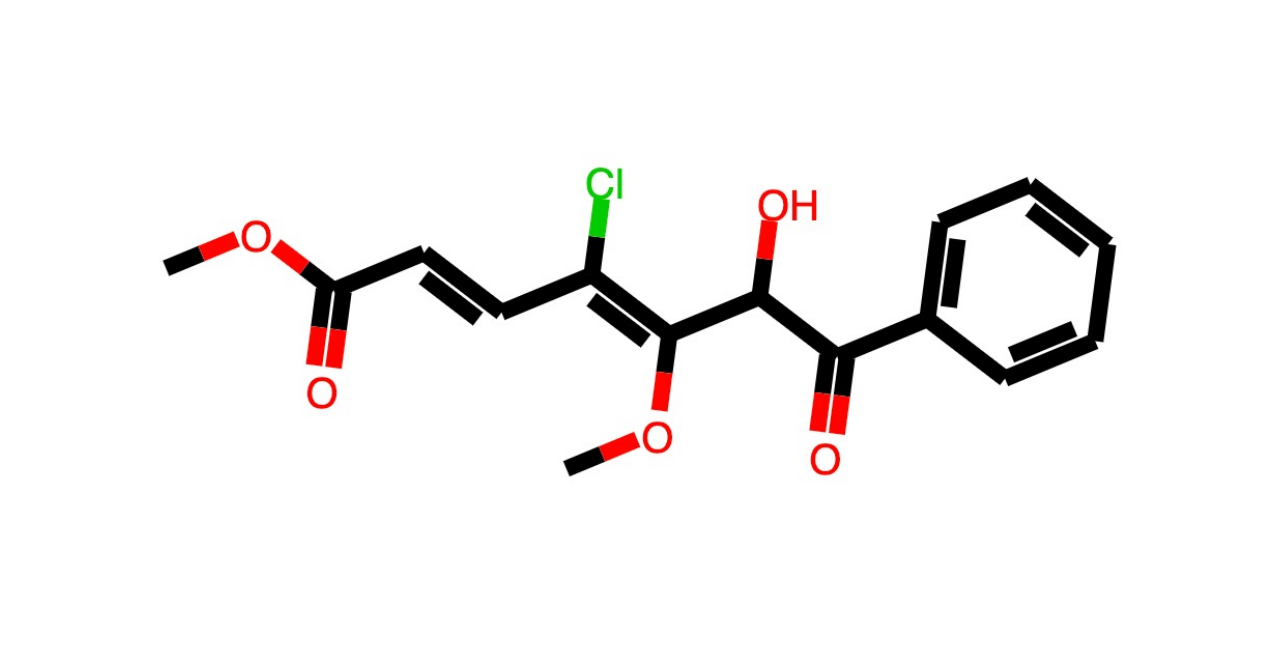} & 
\includegraphics[height=0.66in,valign=c]{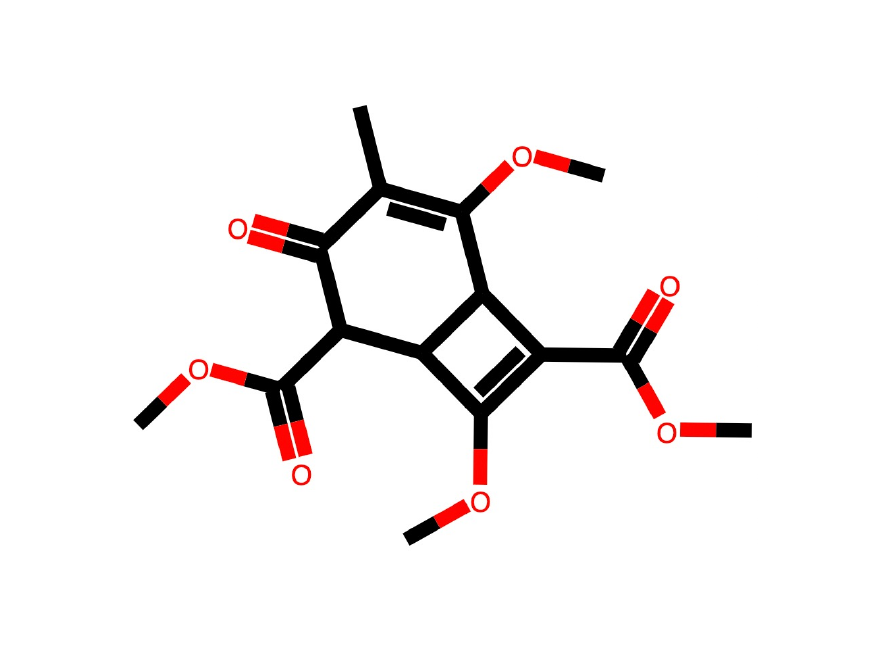} & 
\includegraphics[height=0.6in,valign=c]{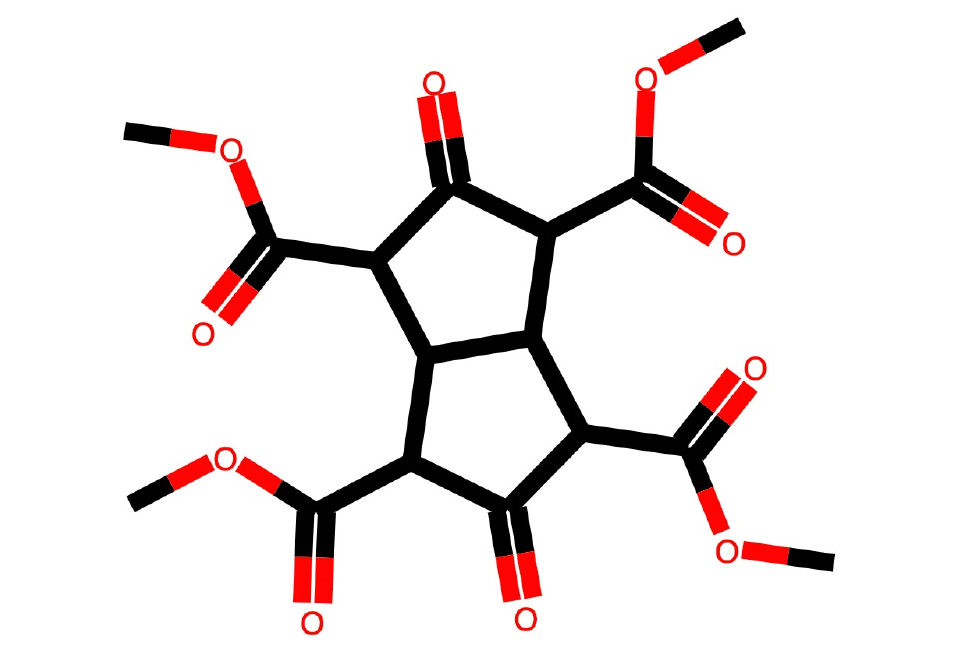}         \\
&  0.154   & 0.146  & 0.157 & \textbf{0.326}  &            \\ \midrule
\multirow{2.5}{*}{BBBP} &  
\includegraphics[height=0.6in,valign=c]{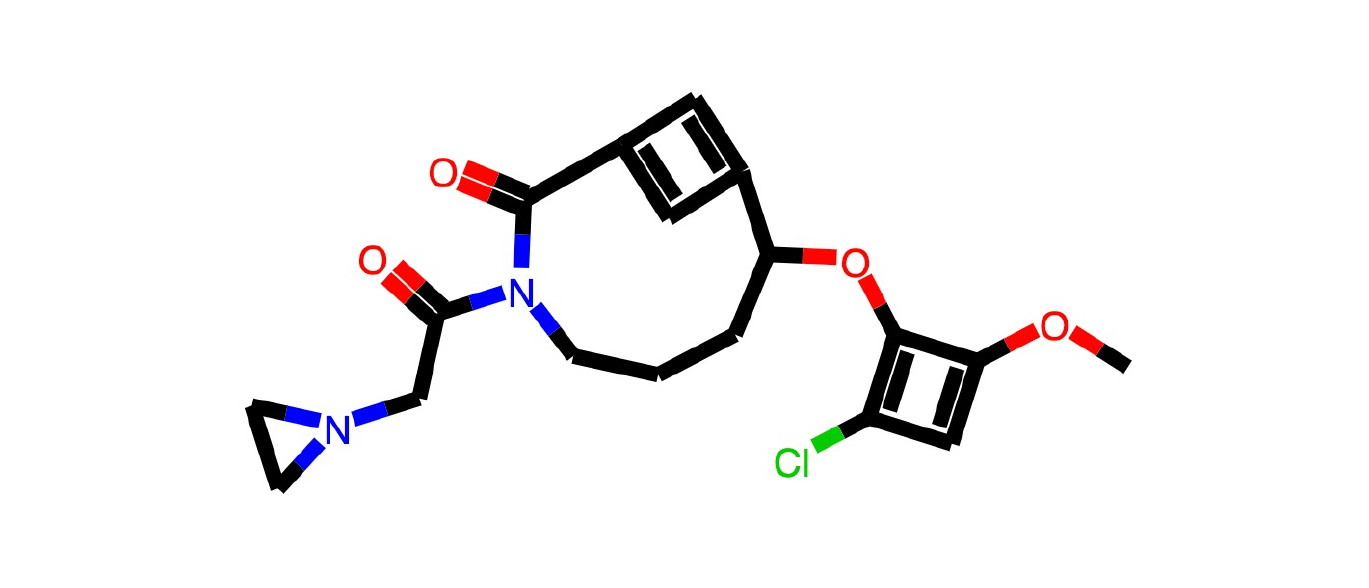} & 
\includegraphics[height=0.6in,valign=c]{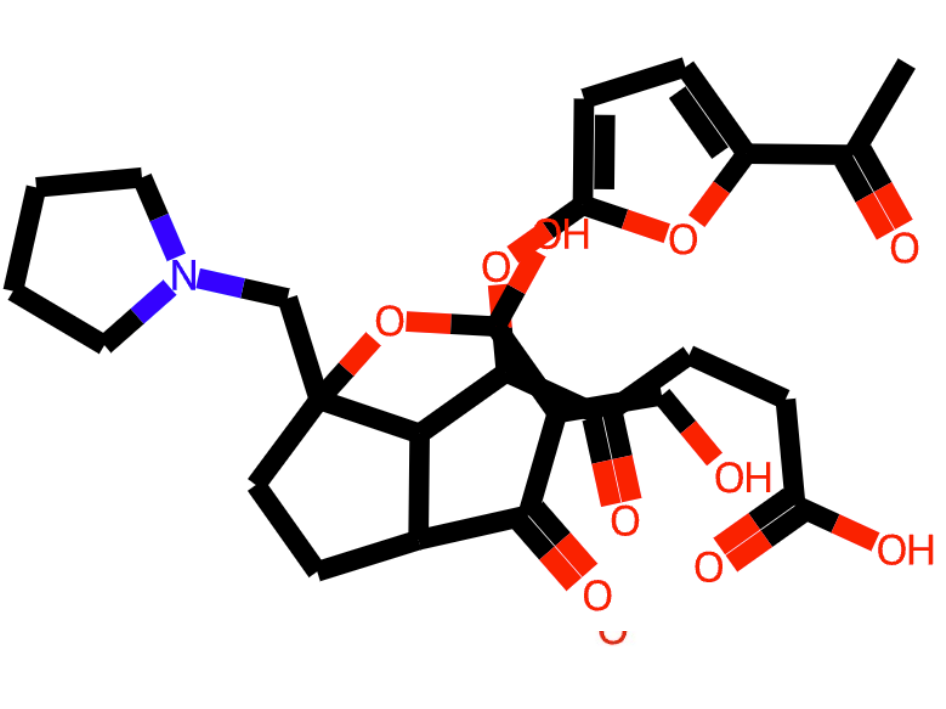} & 
\includegraphics[height=0.6in,valign=c]{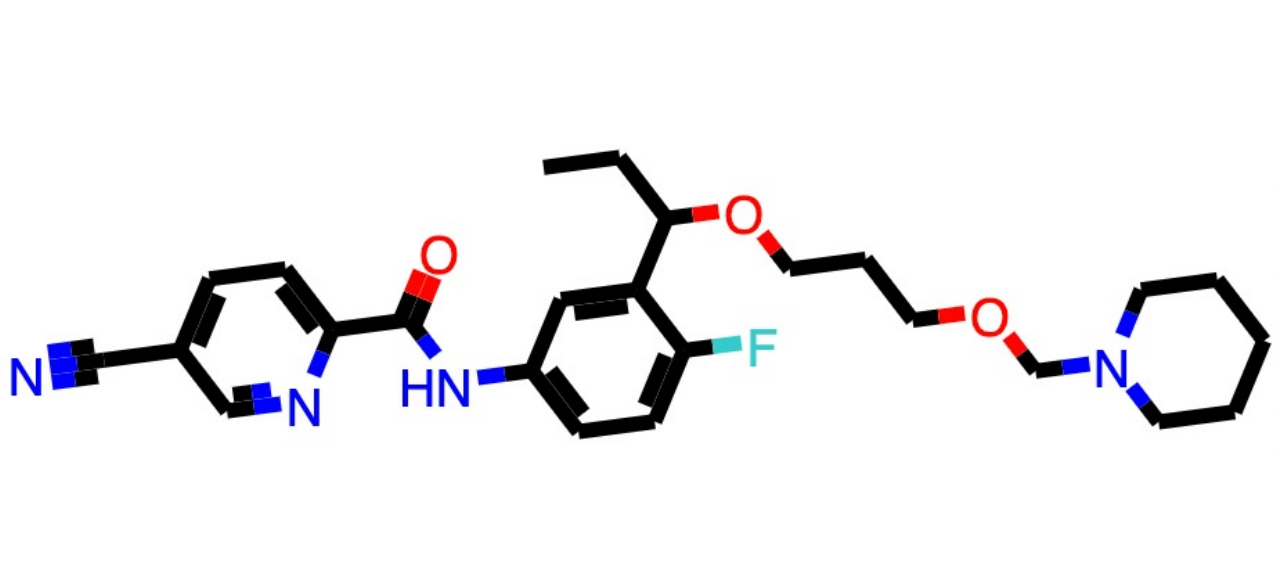} &
\includegraphics[height=0.6in,valign=c]{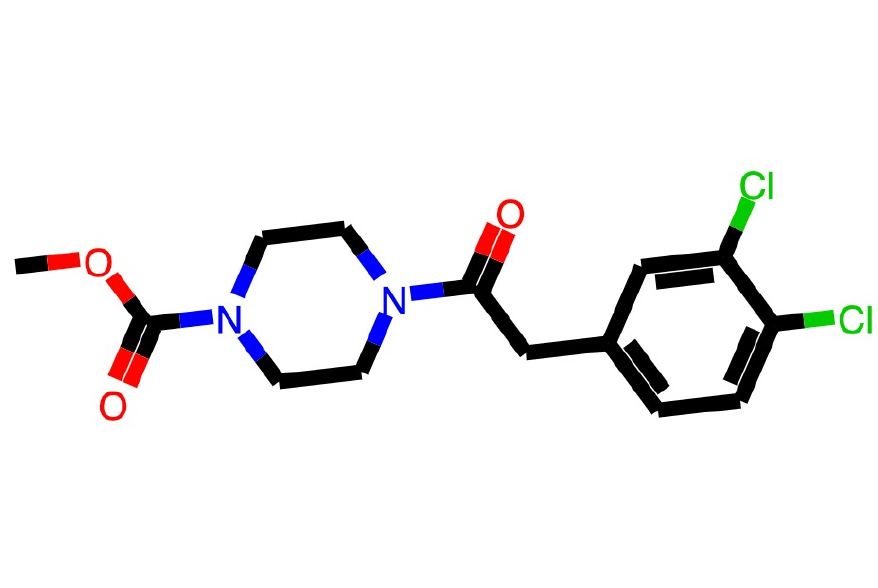} 
& 
\includegraphics[height=0.6in,valign=c]{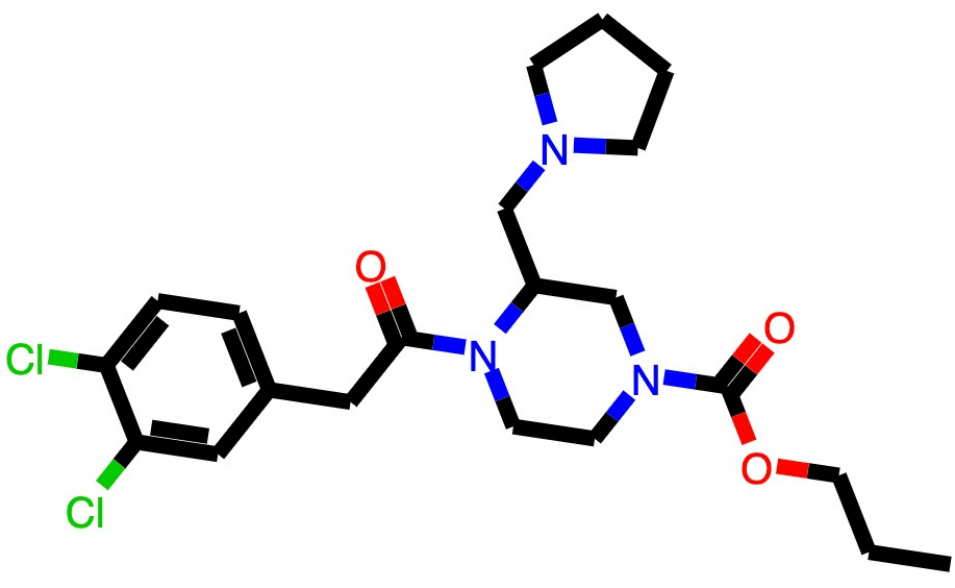} \\
&  0.238   & 0.247  & 0.246 & \textbf{0.505}           \\ \bottomrule
\end{tabular}}
\end{center}
\end{table*}

\section{Experiments}
{We extensively verify the superiority of \Algname by considering various data-efficient molecular generation scenarios. In Section~\ref{subsec:exp_setup}, we explain our experimental setup, e.g., datasets and metrics. In Section~\ref{subsec:main_results}, we present our main molecular generation results on MoleculeNet and QM9 as well as the applicability of our generated molecules in the low-shot molecular property prediction tasks. In Section~\ref{subsec:abla}, we conduct some analysis and an ablation study to validate the effect of each component of our method. We present the application of \Algname on molecular optimization in Appendix~\ref{sup:offline_optimization}. We provide further ablation study and additional experimental results in Appendix~\ref{sup:abla} and \ref{sup:additiona_experiments}, respectively.

\subsection{Experimental Setup}
\label{subsec:exp_setup}

\textbf{Datasets.}
Due to the lack of benchmarks designed particularly for data-efficient molecular generation, we propose to use the following datasets for evaluating molecular generation methods under our problem setup. First, we consider three datasets in the MoleculeNet \citep{wu2018moleculenet} benchmark (originally designed for activity detection): HIV, BBBP, and BACE, which have a significantly smaller number of molecules than popular molecular generation benchmarks \citep{sterling2015zinc,polykovskiy2020molecular}. For example, BACE includes only 691 active molecules. Using the active molecules in each dataset, we construct tasks to generate novel molecules that share the chemical concept, e.g., blood-brain membrane permeability for BBBP. We also utilize these datasets to evaluate the quality of the generated molecules in low-shot molecular property prediction tasks.

Moreover, we utilize the QM9 dataset~\citep{ramakrishnan2014quantum} for our experiments to show the data-efficiency of \Algname. This dataset consists of more than 100k molecules, and thus has become a popular benchmark to evaluate large-scale molecular generation frameworks. Here, we train our method with an extremely small subset of the entire QM9 training split, e.g., 2\% and 10\%, whereas other baseline methods are trained on the entire training split. We provide more details about the datasets in Appendix~\ref{appen:dataset}.

\begin{table*}[t]
\caption{Quantitative results of the generated molecules on the QM9 dataset \citep{ramakrishnan2014quantum}. We mark in Grammar if the method explicitly exploits the grammar of molecular data and thus yields a high Valid. score. Following the setup of \citet{jo2022score}, we report the results using 10,000 sampled molecules. We denote the scores drawn from \citet{luo2022fast} and \citet{ahn2022spanning} with (*) and ($\dagger$), respectively. We mark (-) when the score is not available in the literature. 
We set the highest score in bold. $\uparrow$ and $\downarrow$ indicate higher and lower values are better, respectively. 
For our method, we report the ratio of the number of samples of the dataset used for training.}
\vspace{0.07in}
\label{tab:qm9}
\small
\begin{center}
\begin{tabular}{lccccccc}
\toprule
{Method}     & {Class} & {Grammar} & FCD $\downarrow$   & NSPDK  $\downarrow$ & Valid. $\uparrow$ & Unique. $\uparrow$ & Novelty $\uparrow$ \\ \midrule

CG-VAE$^\dagger$         \citep{liu2018constrained} &             Graph                  &     \textcolor{SJViolet}{\cmark}                     & 1.852  & -     & \textbf{100}      & 98.6       & 94.3    \\
GraphAF \citep{shi2020graphaf}  &        Graph                &                   \textcolor{SJRed}{\xmark}           & 5.268  & 0.020 & 67       & 94.5      & 88.8   \\
MoFlow     \citep{zang2020moflow} &              Graph           &            \textcolor{SJRed}{\xmark}             & 4.467  & 0.017 & 91.4    & {98.7}      & 94.7   \\
EDP-GNN     \citep{niu2020permutation} &                  Graph                &       \textcolor{SJRed}{\xmark}                         & 2.680  & 0.005 & 47.5    & \textbf{99.3}      & 86.6   \\
GraphDF  \citep{luo2021graphdf}  &         Graph              &  \textcolor{SJRed}{\xmark}                                & 10.82 & 0.063 & 82.7    & 97.6      & \textbf{98.1}   \\

GraphEBM    \citep{liu2021graphebm}                              &             Graph & \textcolor{SJRed}{\xmark}             & 6.143  & 0.030 & 8.22     & 97.8      & {97.0}   \\
GDSS         \citep{jo2022score}                                     &             Graph         &     \textcolor{SJRed}{\xmark}  & 2.900  & {0.003} & 95.7    & 98.5      & 86.3   \\
GSDM$^*$          \citep{luo2022fast}            &             Graph               &               \textcolor{SJRed}{\xmark}          & 2.650  & {0.003} & 99.9     & -          & -       \\ 
STGG$^\dagger$               \citep{ahn2022spanning}          &                SMILES           &  \textcolor{SJViolet}{\cmark}                       &{0.585}  & -     & \textbf{100}      & 95.6       & 69.8    \\
 \midrule
\rowcolor{tablegreen}\textbf{\Algname (Ours; 2\%)}                                   &            SMILES &          \textcolor{SJViolet}{\cmark}    & {0.430}  & \textbf{0.001} &  \textbf{100} &   76.1  &  75.6  \\ 

\rowcolor{tablegreen}\textbf{\Algname (Ours; 10\%) }                                  &            SMILES &          \textcolor{SJViolet}{\cmark}& \textbf{0.398}  & \textbf{0.001} &  \textbf{100} &   {88.3}  &  {73.2}  \\

\bottomrule
\end{tabular}
\end{center}
\end{table*}

\begin{table}[t]

\vspace{-0.12in}
\caption{Average $\Delta$ROC-AUC of the low-shot property prediction tasks in the datasets in the MoleculeNet \citep{wu2018moleculenet} benchmark. The results are averaged over 20 random seeds.
}
\vspace{0.05in}
\small
\label{tab:lowshot}
\begin{center}
\begin{tabular}{clcc}
\toprule
{Dataset}                      & \multicolumn{1}{c}{Method}          & 16-shot                     & 32-shot        
\\ 
                               \midrule
\multirow{4.5}{*}{HIV}          
                               & DiGress \citep{vignac2023digress}        & -2.30         & -2.67      
                               \\
                               & MiCaM \citep{geng2023novo}      & 1.02         & 0.69      
                               \\
                               
                               & STGG \citep{ahn2022spanning}         & 0.53          & -0.47            
                               \\
                               \cmidrule{2-4} 
                               &\cellcolor{tablegreen} \textbf{HI-Mol (Ours)}              &\cellcolor{tablegreen} \textbf{2.35}      & \cellcolor{tablegreen}\textbf{2.16}  
                               \\ 
                               \midrule
\multirow{4.5}{*}{BBBP}          
                               & DiGress \citep{vignac2023digress}        & 1.73          & 0.97  
                               \\
                               & MiCaM \citep{geng2023novo}      & 1.91          & 1.78     
                               \\
                               & STGG \citep{ahn2022spanning}         & 1.85           & 1.76      
                               \\
                               \cmidrule{2-4} 
                               & \cellcolor{tablegreen}\textbf{HI-Mol (Ours)}              & \cellcolor{tablegreen}\textbf{2.73}      & \cellcolor{tablegreen}\textbf{2.64} 
                               \\ 
                               \midrule
\multirow{4.5}{*}{BACE}          
                               
                               & DiGress \citep{vignac2023digress}       & -0.60           & -0.91    
                               \\
                               & MiCaM \citep{geng2023novo}      & -0.65          & -1.11      
                               \\
                               & STGG \citep{ahn2022spanning}         & 2.34           & 2.01        
                               \\
                               \cmidrule{2-4} 
                               & \cellcolor{tablegreen}\textbf{HI-Mol (Ours)}              & \cellcolor{tablegreen}\textbf{3.53}      & \cellcolor{tablegreen}\textbf{3.39}  
                               \\ 
                               \bottomrule

\end{tabular}
\vspace{-0.2in}
\end{center}
\end{table}


\textbf{Evaluation setup.}
We consider six metrics that represent diverse aspects which are critical to the evaluation of the generated molecules, e.g., similarity to the target distribution, uniqueness, and novelty. We incorporate some well-known metrics, such as those used in \citet{jo2022score}, as well as introducing a new metric ``Active ratio'':
\begin{itemize}[topsep=1.0pt,itemsep=1.0pt,leftmargin=5.5mm]
    \item [$\bullet$] \textbf{Active ratio\footnote{For reliable evaluation with our metric, we avoid the overlap between the generated molecules and the training data used for generation methods by ignoring the molecule if it is contained in this dataset. Hence, the Novelty score is 100 for all MoleculeNet experiments since all samples are different from the training set (see Table~\ref{tab:moleculenet} for an example). We provide the detailed description of this metric in Appendix~\ref{appen:metrics}. } (Active.)}: Our proposed metric, measuring the ratio of the valid generated molecules that are active, i.e., satisfying the target concept for each task. 
    \item [$\bullet$] \textbf{Fr\'echet ChemNet Distance} (\textbf{FCD};
    \citealp{preuer2018frechet}): Metric for measuring the distance between the source 
    and the target distribution using pre-trained ChemNet. 
    \item [$\bullet$] \textbf{Neighborhood Subgraph Pairwise Distance Kernel MMD} (\textbf{NSPDK}; \citealp{costa2010fast}): Another metric for measuring the gap between source and the target distributions, based on algorithmic computation using graph-based representations of molecules.
    \item [$\bullet$]  \textbf{Validity (Valid.)}: The ratio of the generated molecules that have the chemically valid structure. 
    \item [$\bullet$] \textbf{Uniqueness (Unique.)}: Diversity of the generated molecules based on the ratio of different samples over total valid molecules earned from the generative model.
    \item [$\bullet$]  \textbf{Novelty}: Fraction of the valid molecules that are not included in the training set. 
\end{itemize}

\textbf{Baselines.} We mainly consider the following recently proposed molecular generation methods for evaluation: GDSS \citep{jo2022score}, DiGress \citep{vignac2023digress}, DEG \citep{guo2022data}, JT-VAE \citep{jin2018junction}, PS-VAE \citep{NEURIPS2022_1160792e}, MiCaM \citep{geng2023novo}, CRNN \citep{segler2018generating}, and STGG \citep{ahn2022spanning}. For evaluation on the QM9 dataset \citep{ramakrishnan2014quantum}, we also consider GraphAF~\citep{shi2020graphaf}, GraphDF~\citep{luo2021graphdf}, MoFlow~\citep{zang2020moflow}, EDP-GNN~\citep{niu2020permutation}, and GraphEBM~\citep{liu2021graphebm}, following the recent works \citep{jo2022score,luo2022fast}. We provide more details of the baselines in Appendix~\ref{appen:baselines}.

\subsection{Main Results}
\label{subsec:main_results}
\textbf{Generation on MoleculeNet.} Table~\ref{tab:moleculenet} summarizes the quantitative results of the generated molecules on the HIV, BBBP, and BACE datasets in the MoleculeNet benchmark \citep{wu2018moleculenet}. Our method consistently outperforms other generation methods in terms of Active ratio, FCD, and NSPDK scores on all three datasets. We note that the improvements in these scores are particularly crucial for the deployment of the molecular generation method. For example, the superior Active ratio of \Algname, e.g., $3.7 \rightarrow 11.4$ on the HIV dataset, indicates that the generated molecules are more likely to exhibit the desired activeness on our target task. 
Our method also significantly improves the FCD metric by 20.2 $\rightarrow$ 19.0 and the NSPDK metric by 0.033 $\rightarrow$ 0.019 on the HIV dataset. These improvements highlight the effectiveness of \Algname in generating more faithful molecules that lie in the target distribution. 
We provide qualitative results in Table~\ref{tab:qual_result} by visualizing some of the generated molecules from each dataset. 
We observe that the molecules generated by \Algname capture several crucial common substructures, e.g., many ester groups, while introducing the novel components, e.g., 4-membered ring, due to our unique hierarchical inversion framework.

{We also propose a simple algorithm to modify the generated invalid SMILES strings by correcting invalid patterns\footnote{For example, we modify the invalid SMILES caused by the unclosed ring, e.g., $\mathtt{C1CCC}\rightarrow \mathtt{CCCCC}$. Please see Appendix~\ref{sup:modification} for the detailed algorithm. We mark in the Grammar column in Table~\ref{tab:moleculenet} and \ref{tab:qm9} when modification is applied for evaluation.} without a computational overhead. 
By applying this modification algorithm, we convert an invalid SMILES string to a valid one that represents a valid molecule, therefore, the Validity score becomes 100 in this case. In particular, the molecules from the modified SMILES further improve the overall metrics, e.g., FCD by 19.0 $\rightarrow$ 16.6 and 11.2 $\rightarrow$ 10.7 in the HIV and the BBBP dataset, respectively. This indicates that the modified SMILES indeed represent molecules from the desired low-shot molecule distribution and further highlights the superior quality of our generated molecules.}

}

\begin{table*}[t]
\caption{Ablation of the components of hierarchical textual inversion on the HIV dataset in the MoleculeNet \citep{wu2018moleculenet} benchmark. 
}
\small
\label{tab:abla_hier}
\begin{center}

\begin{tabular}{cc|c|cccccc}
\toprule

            Inversion & Inverted tokens & Grammar & Active. $\uparrow$ & FCD $\downarrow$  & NSPDK $\downarrow$ & Valid. $\uparrow$ & Unique. $\uparrow$ & Novelty $\uparrow$ \\ \midrule

 \textcolor{SJRed}\xmark & - & \textcolor{SJRed}\xmark & 0.0 & 65.3 & 0.450 & 0.4 & \textbf{100} & \textbf{100} \\ 
   \textcolor{SJViolet}\cmark & $[S^\ast]$ &\textcolor{SJRed}{\xmark} & 0.0 & 64.2 & 0.448 & 0.4 & \textbf{100} & \textbf{100} \\
   \textcolor{SJViolet}\cmark & $[S^\ast][D^\ast]$ &\textcolor{SJRed}{\xmark} &  10.2 & 20.3 & 0.021 & 60.0 & 89.3 & \textbf{100}   \\
   \textcolor{SJViolet}\cmark & $[S^\ast][I^\ast][D^\ast]$ &\textcolor{SJRed}{\xmark} &   \textbf{11.4} & 19.0 & \textbf{0.019} & 60.6 & 94.1 & \textbf{100}  \\
 
   \textcolor{SJViolet}\cmark & $[S^\ast][I^\ast][D^\ast]$ &\textcolor{SJViolet}{\cmark} & \textbf{11.4} & \textbf{16.6} & \textbf{0.019} & \textbf{100} & 95.6 & \textbf{100}   \\

\bottomrule

\end{tabular}
\end{center}
\end{table*}

\textbf{Generation on QM9.} In Table~\ref{tab:qm9}, we report the quantitative results of the generated molecules from each method. 
Here, we train our method with a limited portion of data, e.g., 2\% and 10\%, and then compare the results with the baselines that are trained on the entire dataset. Our model shows strong data-efficiency
: only with a 2\% subset of the training data, our method already outperforms the state-of-the-art baseline, STGG \citep{ahn2022spanning}, by 0.585 $\rightarrow$ 0.430 in FCD. Utilizing a 10\% subset further improves the performance of \Algname, reducing the FCD by 0.430 $\rightarrow$ 0.398. In particular, compared with STGG, \Algname not only improves the FCD score but also shows a better Novelty score, which validates the capability of \Algname to find unseen novel molecules from the desired target distribution.

{For an extensive comparison with the baselines which show high Uniqueness and Novelty scores, e.g., GDSS \citep{jo2022score}, we perform an additional comparison after we adjust the Uniqueness and Novelty scores of our method to 100; this setup allows us to perform a fair comparison in the FCD score with these methods. Here, we adjust the sampling strategy slightly; we ignore the generated molecules which have an overlap with the training molecules and the already generated molecules.
Even in this case, \Algname achieves an FCD of 0.601, which outperforms all these baselines. We provide detailed results and discussion in Appendix~\ref{appen:qm9}.}

\begin{figure}[t]
\centering\small
\includegraphics[width=\linewidth]{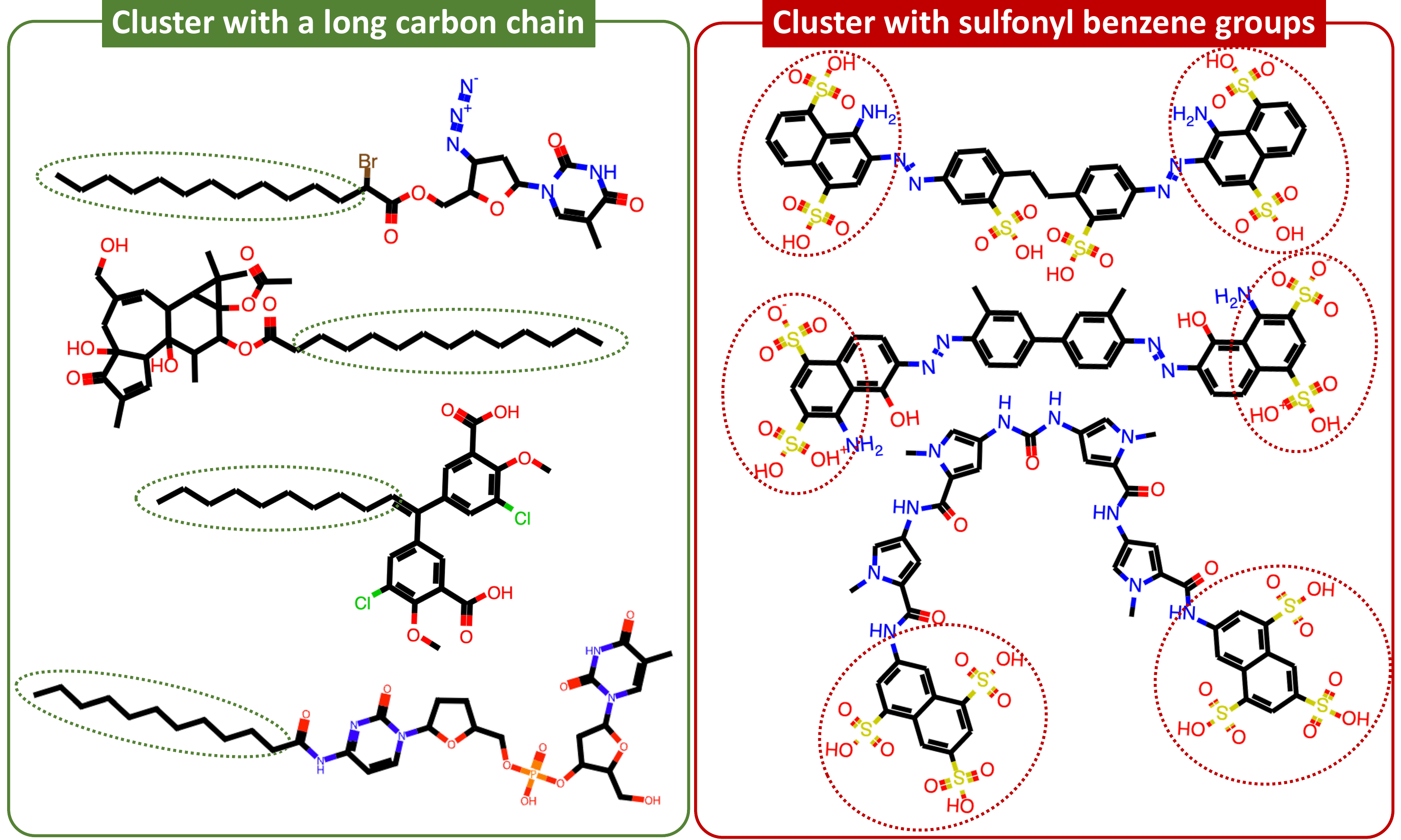}
\vspace{-0.15in}
\caption{
{Visualizations of molecules in two different clusters obtained from the unsupervised clustering objective with the intermediate tokens in Eq. (\ref{eq:training}) on the HIV dataset \citep{wu2018moleculenet}.} 
}
\vspace{-0.15in}
\label{fig:cluster}
\end{figure}

\textbf{Low-shot molecular property prediction.} {We show that the molecules generated by \Algname can be utilized to improve the performance of classifiers for low-shot molecular property prediction. Here, we collect both active and inactive low-shot molecules for each dataset (HIV, BBBP, and BACE) from the MoleculeNet benchmark \citep{wu2018moleculenet}. We separately train molecular generative models for active and inactive molecules, and then generate molecules from the models.
In Table~\ref{tab:lowshot}, we report $\Delta$ROC-AUC scores\footnote{This score is calculated by the improvement in the ROC-AUC score when the generated molecules are additionally added to the original low-shot training data; higher is better.} from each method. We find that \Algname consistently shows the superior $\Delta$ROC-AUC scores in various low-shot property prediction tasks. This demonstrates the efficacy of \Algname to learn the common concept, i.e., activeness and in-activeness, of each molecular property prediction task even with a limited number of molecules. In practical scenarios, where the label information is hard to achieve, our \Algname indeed plays an important role in improving the classifier. We provide experimental details in Appendix~\ref{appen:classification_detail}.}

\textbf{Extremely limited data regime.}
Since our model exploits the power of large molecular language models by designing a molecule-specialized textual inversion scheme, one can expect our model to be beneficial in extremely limited data regimes compared with prior methods. To verify this, we conduct an experiment using only subset of the HIV dataset and report its quantitative result in Table~\ref{tab:moleculenet_few}. Even with this situation, \Algname still outperforms prior state-of-the-art molecular generation methods, e.g., our method improves FCD as 39.2 $\to$ 34.8 when trained with 30 samples.

\subsection{Analysis}
\label{subsec:abla}

\textbf{Effect of intermediate tokens.} 
{Recall that we have introduced intermediate tokens $\{[I^{\ast}_{k}]\}_{k=1}^{K}$ in our hierarchical textual inversion framework, which are selected in an unsupervised manner during the inversion to learn some of the cluster-wise features included in given molecules (see Eq.~(\ref{eq:training})). To validate the effect of our text token design, we visualize the clustering results in Figure~\ref{fig:cluster} by providing groups of the molecules that are assigned to the same intermediate token. As shown in this figure, molecules are well grouped according to their common substructures, e.g., a long carbon chain or sulfonyl benzene groups. 
Such a learning of cluster-wise low-level semantics is indeed beneficial in molecular generation, since molecules often share the same chemical concept, e.g., blood-membrane permeability, even when they have large structural differences. }

\begin{table*}[ht]
\caption{Results of molecular generation on subsets of the HIV dataset \citep{wu2018moleculenet}. 
We generate the same number of molecules as the number of the training samples. 
Due to the large training cost, we report the score of DEG~\citep{guo2022data} only for 30 samples.}
\vspace{-0.1in}
\label{tab:moleculenet_few}
\begin{center}
\small

\resizebox{\textwidth}{!}{
\begin{tabular}{clcccccccc}
\toprule
           \# Samples      &     Method & Class & Grammar &  Active. $\uparrow$ & FCD $\downarrow$ & NSPDK $\downarrow$ & Valid. $\uparrow$ & Unique. $\uparrow$ & Novelty $\uparrow$ \\ \midrule
                        
\multirow{4.5}{*}{30} &DEG \citep{guo2022data}      & Graph           &\textcolor{SJViolet}{\cmark} & \phantom{0}3.3 & {39.2} & {0.105} & \textbf{100}& \textbf{100} & \textbf{100}\\ 
&STGG \citep{ahn2022spanning} & SMILES &\textcolor{SJViolet}{\cmark} & \phantom{0}0.0 & 41.5 & 0.110 & \textbf{100} & 67 & \textbf{100}   \\ 
&CRNN \citep{segler2018generating} & SMILES &\textcolor{SJRed}{\xmark} & \phantom{0}0.0 & 40.0 & 0.121 & 80 & 71 & \textbf{100}   \\
 \cmidrule{2-10}
&\cellcolor{tablegreen}\textbf{\Algname (Ours)} & \cellcolor{tablegreen}SMILES &\cellcolor{tablegreen}\textcolor{SJRed}{\xmark} & \cellcolor{tablegreen}\phantom{0}\textbf{8.3} &    \cellcolor{tablegreen}\textbf{34.8}& \cellcolor{tablegreen}\textbf{0.103} & \cellcolor{tablegreen}80  & \cellcolor{tablegreen}{75} & \cellcolor{tablegreen}\textbf{100}  \\

\midrule

\multirow{3.5}{*}{150} & STGG \citep{ahn2022spanning} & SMILES &\textcolor{SJViolet}{\cmark} & \phantom{0}1.3 & 28.2 & 0.054 & \textbf{100} & 90 & \textbf{100}   \\ 
&CRNN \citep{segler2018generating} & SMILES &\textcolor{SJRed}{\xmark}  &\phantom{0}1.3 & 30.1 & 0.063 & 50 & 84  & \textbf{100} \\
 \cmidrule{2-10}

&\cellcolor{tablegreen}\textbf{\Algname (Ours)} & \cellcolor{tablegreen}SMILES &\cellcolor{tablegreen}\textcolor{SJRed}{\xmark} &   \cellcolor{tablegreen}\phantom{0}\textbf{8.3} & \cellcolor{tablegreen}\textbf{22.1}&\cellcolor{tablegreen} \textbf{0.038} &\cellcolor{tablegreen}64  & \cellcolor{tablegreen}{\textbf{91}} & \cellcolor{tablegreen}\textbf{100}  \\ \midrule

\multirow{3.5}{*}{500} & STGG \citep{ahn2022spanning} & SMILES &\textcolor{SJViolet}{\cmark} & \phantom{0}1.3 & 22.8 & 0.041 &\textbf{100} & 74 & \textbf{100}  \\ 
&CRNN \citep{segler2018generating} & SMILES &\textcolor{SJRed}{\xmark} & \phantom{0}2.7 & 30.0 & 0.064 & 51 & \textbf{100} & \textbf{100}\\
 \cmidrule{2-10}

&\cellcolor{tablegreen}\textbf{\Algname (Ours)} &\cellcolor{tablegreen} SMILES &\cellcolor{tablegreen}\textcolor{SJRed}{\xmark} &  \cellcolor{tablegreen}\textbf{10.3}  &\cellcolor{tablegreen}\textbf{20.8} & \cellcolor{tablegreen}\textbf{0.020}&\cellcolor{tablegreen} 63 &\cellcolor{tablegreen}91&\cellcolor{tablegreen}\textbf{100}\\

\bottomrule

\end{tabular}
}
\end{center}
\end{table*}

\textbf{Ablation on hierarchical textual inversion.} 
{To validate the effectiveness of each component in our \Algname framework, we 
compare the results 
where some 
components are excluded from the overall framework. Specifically, we compare the generation performance of the following setups: (1) not using the inversion technique; we train the text-to-molecule model with the molecule-description pairs, (2) using the shared token $[S^\ast]$ only, (3) using $[S^\ast]$ and the detail tokens $[D_n^{\ast}]$, (4) using all three types of tokens, and (5) applying the additional modification algorithm. Note that for (1) and (2), it is impossible to apply our interpolation-based sampling; instead, 
we use temperature sampling 
with temperature $\tau=2.0$. We provide this result in Table~\ref{tab:abla_hier}. First, we find that (1) the na\"ive training and (2) the inversion with a single shared token \citep{gal2022image} do not show reasonable performance, i.e., they achieve only 0.4\% Validity. 
In (3) and (4), introducing low-level tokens in the inversion framework significantly improves the generation quality by learning the low-level features in molecules. 
Finally, (5) the modification algorithm converts an invalid generated SMILES into a valid one that lies in our target distribution. We provide additional ablation results in Appendix \ref{sup:abla}.
}

\section{Conclusion}
We propose \Algname, a data-efficient molecular generation framework that utilizes a molecule-specialized textual inversion scheme. Specifically, we propose to capture the hierarchical information of molecular data in the inversion stage, and use it to sample novel molecules. 
We hope our method initiates under-explored but crucial research direction in the data-efficient generation of molecules.


\textbf{Limitation and future work.} In this work, we apply our novel hierarchical textual inversion scheme to the molecular language model \citep{edwards2022translation}, where developing such a model is a very recently considered research direction. An important future work would be improving the large-scale molecular language models themselves, e.g., the breakthroughs in the image domain \citep{rombach2022high}, which will allow more intriguing applications of our \Algname framework, such as composition (see Appendix \ref{sup:additiona_experiments}). 

\section*{Impact Statement}
This work will facilitate research in molecular generation, which can speed up the development of 
many 
important generation tasks such as finding drugs for a specific organ and disease when the hit molecules are rarely known.
However, malicious use of well-learned molecular generative model poses a potential threat of creating hazardous molecules, such as toxic chemical substances. It is an important research direction to prevent malicious usages of generative models \citep{openai2023gpt4}.
On the other hand, molecular generation is also essential for generating molecules to defend against harmful substances, so the careful use of our work, \Algname, can lead to more positive effects. 

\section*{Acknowledgements}
We thank all the anonymous reviewers for their insightful feedbacks and discussions. This work was supported by Institute of Information \& communications Technology Planning \& Evaluation (IITP) grant funded by the Korea government (MSIT) (No.RS-2019-II190075, Artificial Intelligence Graduate School Program(KAIST), No.RS-2021-II212068, Artificial Intelligence Innovation Hub) and Mogam institute for Biomedical Research and GC Biopharma.

\nocite{langley00}

\bibliography{reference}
\bibliographystyle{icml2024}

\newpage
\appendix
\onecolumn
\newpage
\begin{center}
{\bf {\Large Appendix: Data-Efficient Molecular Generation \\ with Hierarchical Textual Inversion}} 
\end{center}
\vspace{0.07in}

\section{Method Details}
\label{appen:method}

We utilize a recently introduced text-to-molecule model, MolT5-Large-Caption2Smiles \citep{edwards2022translation} in our \Algname framework.\footnote{\url{https://huggingface.co/laituan245/molt5-large-caption2smiles}} This model is constructed upon a text-to-text model, T5 \citep{raffel2020exploring}, and molecular information is injected by additional training with both unpaired SMILES \citep{weininger1988smiles} string and caption-SMILES paired dataset. We update the token embeddings and the linear heads, while freezing other parameters. Our experiment is conducted for 1,000 epochs using a single NVIDIA GeForce RTX 3090 GPU with a batch size of 4. We use AdamW optimizer with $\epsilon=1.0\times 10^{-8}$ and let the learning rate $0.3$ with linear scheduler. We clip gradients with the maximum norm of 1.0. We update the assigned cluster $c_n$ of each molecule for the first 5 epochs following Eq. (\ref{eq:training}). For interpolation-based sampling, we choose a uniform distribution $p(\lambda)$, (i.e., $p(\lambda) \coloneqq \mathcal{U}(l,1-l)$), where $\lambda$ controls relative contributions of interpolated token embeddings. We set $l=0.0$ on the datasets in MoleculeNet benchmark \citep{wu2018moleculenet}, and $l=0.3$ on the QM9 dataset \citep{ramakrishnan2014quantum}.

\section{Datasets}
\label{appen:dataset}

\textbf{MoleculeNet dataset.} We perform generation experiments on single-task datasets, HIV, BBBP, and BACE, from MoleculeNet \citep{wu2018moleculenet} benchmark. For each dataset, molecules are labeled with 0 or 1, based on its activeness of the target property:

\begin{itemize}[topsep=1.0pt,itemsep=1.0pt,leftmargin=4.5mm]
    \item [$\bullet$] \emph{HIV} consists of molecules and its capability to prevent HIV replication.
    \item [$\bullet$] \emph{BBBP} consists of molecules and whether each compound is permeable to the blood-brain barrier.
    \item [$\bullet$] \emph{BACE} consists of molecules and its binding results for a set of inhibitors of $\beta$-secretase-1.
\end{itemize}

We collect active (e.g., label-1) molecules to train molecular generative models. We utilize a common splitting scheme for MoleculeNet dataset, \emph{scaffold split} with split ratio of train:valid:test = 80:10:10 \citep{wu2018moleculenet}. We emphasize that such \emph{scaffold split} is widely considered in molecular generation domain \citep{ahn2022spanning}. Additional statistics for datasets on MoleculeNet are provided in Table~\ref{tab:moleculenet_stat}.

\begin{table*}[h]
\caption{MoleculeNet downstream classification dataset statistics}
\begin{center}
\vspace{0.1in}
\label{tab:moleculenet_stat}
\begin{tabular}{l|ccccccccc}
    \toprule
    
    Dataset & HIV & BBBP & BACE\\ 
    \midrule
    Number of molecules & 41,127 &  2,039 & 1,513\\
    Number of active molecules & 1,443 & 1,567 &691 \\
    Avg. Node & 25.51 & 24.06 & 34.08\\
    Avg. Degree & 54.93 & 51.90 & 73.71\\
    \bottomrule
\end{tabular}

\end{center}
\end{table*}

\textbf{QM9 dataset.} We perform generation experiments on the QM9 dataset \citep{ramakrishnan2014quantum}, which is a widely adopted to benchmark molecular generation methods. This dataset consists of 133,885 small orginic molecules. We follow the dataset splitting scheme of \citep{ahn2022spanning} and randomly subset the training split with 2\%, 5\%, 10\%, 20\% and 50\% ratio for training our \Algname.

\newpage
\section{Evaluation Metrics}
\label{appen:metrics}
We mainly utilize 6 metrics to incorporate diverse aspects for evaluation of the generated molecules. We adopt 5 metrics (FCD, NSPDK, Validity, Uniqueness, Novelty) used in \citep{jo2022score}: 

\begin{itemize}[topsep=1.0pt,itemsep=1.0pt,leftmargin=5.5mm]
    \item [$\bullet$] \textbf{Fr\'echet ChemNet Distance (FCD)} \citep{preuer2018frechet} evaluates the distance between the generated molecules and test molecules using the activations of the penultimate layer of the ChemNet, similar to popular Fr\'echet inception distance (FI) used in image domain~\citep{heusel2017gans}:
    \begin{align}
    \label{eq:fcd}
    \texttt{FCD}\coloneqq \lVert m-m_g \rVert_2^2 + \text{Tr}\big(C + C_g - 2(CC_g)^{1/2}\big),
    \end{align}
    where $m, C$ are the mean and covariance of the activations of the test molecules, and $m_g, C_g$ are the mean and covariance of the activations of the generated molecules. 
    \item [$\bullet$] \textbf{Neighborhood Subgraph Pairwise Distance Kernel MMD (NSPDK)} \citep{costa2010fast} calculates the maximum mean discrepancy between the generated molecules and test molecules. We follow the evaluation protocol in \citep{jo2022score}, to incorporate both atom and bond features.
    \item [$\bullet$]  \textbf{Validity (Valid.)} is the ratio of the generated molecules that does not violate chemical validity, e.g., molecules that obey the valency rule.
    \item [$\bullet$] \textbf{Uniqueness (Unique.)} is the ratio of different samples over total valid generated molecules.
    \item [$\bullet$]  \textbf{Novelty} is the ratio of valid generated molecules that are not included in the training set. 
\end{itemize}

We introduce an additional metric (Active ratio) to evaluate how the generated molecules are likely to be active, e.g., label-1 on our target property:
\begin{itemize}[topsep=1.0pt,itemsep=1.0pt,leftmargin=5.5mm]
    \item [$\bullet$] \textbf{Active ratio (Active.)} is the ratio of the valid generated molecules that are active. 
\end{itemize}
We utilize pre-trained classifiers to measure the activeness of the generated molecules. To be specific, we train a graph isomorphism network \citep[GIN,][]{xu2018how} with the entire training split, e.g., contains both active (label-1) and inactive (label-0) molecules, of each dataset in the MoleculeNet benchmark \citep{wu2018moleculenet}. We train 5-layer GIN with a linear projection layer for 100 epochs with Adam optimizer, a batch size of 256, a learning rate of 0.001, and a dropout ratio of 0.5. We select the classifier of the epoch with the best validation accuracy. The accuracies of the pre-trained classifier on the validation split are 98.2\%, 86.3\%, and 86.1\%, respectively. We calculate Active ratio by the ratio of the generated molecules that this classifier classifies as label-1. 

\newpage
\section{Baselines}
\label{appen:baselines}
In this paper, we compare our method with an extensive list of baseline methods in the literature of molecular generation.
We provide detailed descriptions of the baselines we considered:

\begin{itemize}[topsep=1.0pt,itemsep=1.0pt,leftmargin=5.5mm]
    \item [$\bullet$]  \textbf{GDSS} \citep{jo2022score} proposes a diffusion model for graph structure, jointly learning both node and adjacency space by regarding each attributes as continuous values.
    \item [$\bullet$]  \textbf{DiGress} \citep{vignac2023digress} proposes a discrete diffusion process for graph structure to properly consider categorical distributions of node and edge attributes.
    \item [$\bullet$]  \textbf{DEG} \citep{guo2022data} suggests constructing molecular grammars from automatically learned production rules for data-efficient generation of molecules. Due to the high computational complexity of the grammar construction, this method can only be applied to structurally similar molecules, e.g., monomers or chain-extenders, with an extremely limited number of molecules ($\sim$100 molecules with high structural similarity). Nevertheless, we compare with this method in the extremely limited data regime of Appendix~\ref{sup:additiona_experiments}.
    \item [$\bullet$] \textbf{JT-VAE} \citep{jin2018junction} proposes a variational auto-encoder that represents molecules as junction trees, regarding motifs of molecules as the nodes of junction trees. 
    \item [$\bullet$] \textbf{PS-VAE} \citep{NEURIPS2022_1160792e} utilizes a principal subgraph as a building block of molecules and generates molecules via merge-and-update subgraph extraction.
    \item [$\bullet$] \textbf{MiCaM} \citep{geng2023novo} introduces a connection-aware motif mining method to model the target distribution with the automatically discovered motifs.
    \item [$\bullet$] \textbf{CRNN} \citep{segler2018generating} builds generative models of SMILES strings with recurrent decoders.
    \item [$\bullet$]  \textbf{STGG} \citep{ahn2022spanning} introduces a spanning tree-based molecule generation which learns the distribution of intermediate molecular graph structure with tree-constructive grammar.
    \item [$\bullet$]  \textbf{GraphAF} \citep{shi2020graphaf} proposes an auto-regressive flow-based model for graph generation.
    \item [$\bullet$]  \textbf{GraphDF} \citep{luo2021graphdf} introduces an auto-regressive flow-based model with discrete latent variables.
    \item [$\bullet$]  \textbf{MoFlow} \citep{zang2020moflow} utilizes a flow-based model for one-shot molecular generation.
    \item [$\bullet$]  \textbf{EDP-GNN} \citep{niu2020permutation} proposes a one-shot score-based molecular generative model, utilizing a discrete-step perturbation procedure of node and edge attributes.
    \item [$\bullet$]  \textbf{GraphEBM} \citep{liu2021graphebm} introduces a one-shot energy-based model to generate molecules by minimizing energies with Langevin dynamics.
    \item [$\bullet$]  \textbf{GSDM} \citep{luo2022fast} is a follow-up work of GDSS \citep{jo2022score}, suggesting to consider the spectral values of adjacency matrix instead of adjacency matrix itself.
    \item [$\bullet$]  \textbf{CG-VAE} \citep{liu2018constrained} proposes a recursive molecular generation framework that generates molecules satisfying the valency rules by masking out the action space.
    
\end{itemize}

\newpage
\section{Results with Other Molecular Language Models}
\begin{table}[h]
\caption{Quantitative results of generated molecules with \Algname varying the molecular language models.}
\vspace{0.1in}
\label{tab:ablation_model}
\small
\begin{center}
\begin{tabular}{c|c|cccccc}
\toprule

           Dataset & Model &  Active. $\uparrow$ & FCD $\downarrow$  & NSPDK $\downarrow$ & Valid. $\uparrow$ & Unique. $\uparrow$ & Novelty $\uparrow$ \\ \midrule

 \multirow{3}{*}{HIV}& MolT5 \citep{edwards2022translation} & \textbf{11.4} & \textbf{16.6} & \textbf{0.019} & \textbf{100} & 95.6 & \textbf{100}   \\ 
 &  ChemT5 \citep{christofidellis2023unifying} & 10.8 & 16.8 & \textbf{0.019} & \textbf{100} & 98.6 & \textbf{100} 
 \\
 &  BioT5 \citep{pei2023biot5} & 10.1 & 16.9 & 0.023  & \textbf{100} & \textbf{99.4} & \textbf{100} 
 \\ \midrule
 
 \multirow{3}{*}{BBBP}& MolT5 \citep{edwards2022translation} & \textbf{94.6} & \textbf{10.7} & \textbf{0.009} & \textbf{100} & 94.2 & \textbf{100}  \\ 
 & ChemT5 \citep{christofidellis2023unifying} & 93.2 & 10.8 & 0.011 & \textbf{100} & 96.6 & \textbf{100} 
 \\
 &  BioT5 \citep{pei2023biot5} & 92.8 & 11.4 & 0.013 & \textbf{100} & \textbf{99.0} & \textbf{100}  
 \\ \midrule
 
   \multirow{3}{*}{BACE}& MolT5 \citep{edwards2022translation} & {80.4} & {14.0} & {0.039} & \textbf{100} & 74.4 & \textbf{100}  \\ 
 &  ChemT5 \citep{christofidellis2023unifying} & \textbf{83.1} & \textbf{13.6} & \textbf{0.036} & \textbf{100} & 87.0 & \textbf{100} \\
 &  BioT5 \citep{pei2023biot5} & 82.4 & 14.3 & 0.038 & \textbf{100} & \textbf{98.6} & \textbf{100}  \\

\bottomrule

\end{tabular}
\end{center}
\end{table}
\label{sup:model_ablation}
{In Table~\ref{tab:ablation_model}, we show the experimental results of our \Algname framework based on varying molecular language models. We utilize Large-Caption2Smiles \citep[MolT5,][]{edwards2022translation}, Text+Chem T5-augm \citep[ChemT5,][]{christofidellis2023unifying}, and BioT5 \citep[BioT5,][]{pei2023biot5}. The results show that the performance of \Algname is consistent across various molecular language models, i.e., \Algname framework reliably generates high quality molecules that lie in the desired target distribution. }

\section{Offline Molecular Property Optimization}
\label{sup:offline_optimization}
    \begin{figure}[ht]

\begin{minipage}{0.5\textwidth}
\captionof{table}{Results of molecular property maximization task. We report the top-3 property scores denoted by \texttt{1st}, \texttt{2nd}, and \texttt{3rd}. The baseline scores are drawn from \citet{ahn2022spanning}. 
}
\vspace{0.1in}
\label{tab:optimization}
    
    \resizebox{\textwidth}{!}{%
    \begin{tabular}{lcccccc}
    
\toprule
& & &\multicolumn{3}{c}{{PlogP}}\\\cmidrule{3-6}
Method & Class & Offline & \texttt{1st} & \texttt{2nd} & \texttt{3rd} \\ \midrule
GVAE \citep{kusner2017grammar} & SMILES & \cmark &2.94 & 2.89 & 2.80 \\
SD-VAE \citep{dai2018syntax}   &  Syntax Tree & \cmark &4.04 & 3.50 & 2.96 \\
JT-VAE \citep{jin2018junction} & Fragment& \xmark &5.30 & 4.93 & 4.49 \\
MHG-VAE \citep{kajino2019molecular} & Fragment & \xmark &5.56 & 5.40 & 5.34 \\
GraphAF \citep{shi2020graphaf} & Graph& \xmark &12.23 & 11.29 & 11.05 \\
GraphDF \citep{luo2021graphdf} & Graph& \xmark &13.70 & 13.18 & 13.17 \\
STGG \citep{ahn2022spanning} & SMILES & \cmark &23.32 & 18.75 & 16.50 \\

 \midrule

\rowcolor{tablegreen}\textbf{\Algname (Ours; 1\%)}   &SMILES & \cmark& \textbf{24.67}  & \textbf{21.72}  & \textbf{20.73}\\ 

\bottomrule
\end{tabular}}
    \end{minipage}\hfill
\begin{minipage}{0.45\textwidth}\centering
\includegraphics[width=\linewidth]{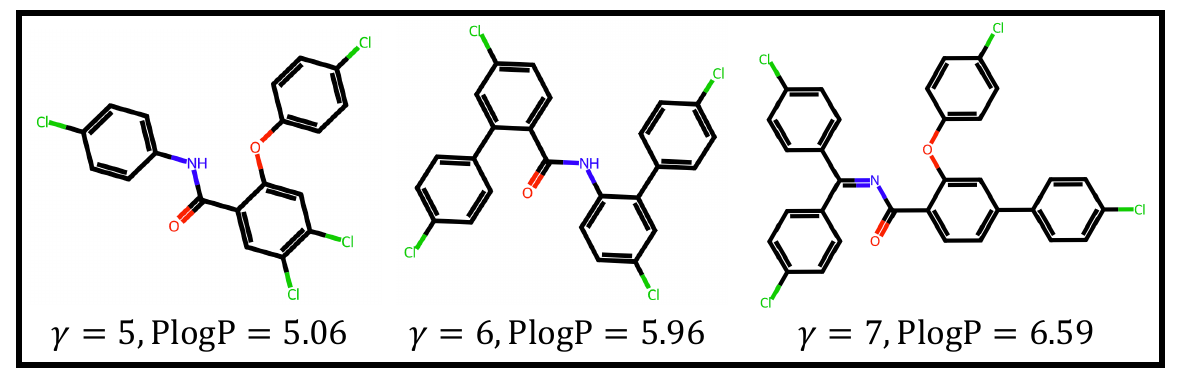} 
\vspace{-0.25in}

\caption{Visualization of the generated molecules with the specific condition $\gamma$. The maximum PLogP among the training molecules is 4.52.}
\label{fig:conditional_generation}
    
\end{minipage}

    \end{figure}
{In this section, we show the applicability of our \Algname in molecular optimization, mainly following the experimental setup of \citet{ahn2022spanning}. Specifically, we consider the offline\footnote{While some online optimization algorithms show promising performances \citep{jensen2019graph,fu2021differentiable}, they require the specific value of the relevant property for the intermediate molecules in the learning process. This additional cost often limits the practical application of online algorithms since the calculation of the property sometimes requires high experimental costs \citep{gao2022sample}.} molecular property optimization task on the penalized octanol-water partition coefficient (PlogP). We train a conditional molecular generative model $p_{\mathrm{model}}(\mathbf{x}|\gamma)$ under the \Algname framework where $\gamma$ denotes the PlogP value. Then, we sample with a high $\gamma$ to generate molecules with high PLogP. In Table~\ref{tab:optimization}, our \Algname generates molecules with high PLogP even when trained with only 1\% of the entire training dataset. Here, we remark that solely maximizing the molecular property (such as PLogP) may generate unrealistic molecules \citep{ahn2022spanning}, e.g., unstable or hard-to-synthesize (see Appendix~\ref{sup:optimization}). To address this and highlight the practical application of our \Algname framework, we further show the model's capability to generate molecules with the desired PLogP. In Figure~\ref{fig:conditional_generation}, \Algname generates realistic molecules with the target PLogP, even when the desired condition $\gamma$ is unseen in the training molecules. The overall results show that our \Algname exhibits a huge potential for real-world scenarios where we aim to generate molecules with a specific target property.}

\newpage
\section{Ablation Study}
\label{sup:abla}

\begin{table}[h]
\caption{Ablation on the text prompts for interpolation-based sampling on the 2\% subset of QM9.
}
\vspace{0.1in}
\begin{center}\small
\label{tab:prompt_ablation}
\begin{tabular}{lccccc}
\toprule
{Generation prompt}     & FCD $\downarrow$   & NSPDK $\downarrow$ & Valid. $\uparrow$ & Unique. $\uparrow$ & Novelty $\uparrow$ \\ 
\midrule
${\mathsf{The\,\,molecule\,\,is\,\,a\,\,} [S^\ast][I^\ast_{c_n}][D^\ast_n]}$& \textbf{0.210}  & \textbf{0.001} &  \textbf{92.2} &   {61.4}  & 47.5 \\
\midrule

${\mathsf{The\,\,molecule\,\,is\,\,similar\,\,to\,\,} [S^\ast][I^\ast_{c_n}][D^\ast_n]}$& 0.234 & \textbf{0.001}& 91.1& 63.4&50.6\\
${\mathsf{A\,\,similar\,\,molecule\,\,of\,\,} [S^\ast][I^\ast_{c_n}][D^\ast_n]}$& 0.271 & \textbf{0.001}& 91.5 & 65.0& 52.6 \\
${\mathsf{The\,\,chemical\,\,is\,\,similar\,\,to\,\,} [S^\ast][I^\ast_{c_n}][D^\ast_n]}$&  0.437 & 0.002 & 90.2 & 75.5 & 72.4\\ \midrule
${\mathsf{A\,\,similar\,\,chemical\,\,of\,\,} [S^\ast][I^\ast_{c_n}][D^\ast_n]}$& {0.434}  & \textbf{0.001} &  90.7 &   \textbf{75.8}  &  \textbf{73.5}  \\ \bottomrule
\end{tabular}

\vspace{-0.3in}
\end{center}
\end{table}
\begin{table*}[ht]
\caption{Ablation of hierarchical textual inversion on the HIV dataset in the MoleculeNet \citep{wu2018moleculenet} benchmark. 
}
\vspace{0.1in}
\small
\label{tab:abla_hier_sup}
\begin{center}

\begin{tabular}{c|c|ccccc}
\toprule

             Inverted tokens & Active. $\uparrow$ & FCD $\downarrow$  & NSPDK $\downarrow$ & Valid. $\uparrow$ & Unique. $\uparrow$ & Novelty $\uparrow$ \\ \midrule

    $[D^\ast]$  & 5.4 & 21.7 & 0.026 & \textbf{100} & 88.8 & \textbf{100} \\ \midrule
    $[S^\ast][I^\ast][D^\ast]$  & \textbf{11.4} & \textbf{16.6} & \textbf{0.019} & \textbf{100} & \textbf{95.6} & \textbf{100}   \\

\bottomrule

\end{tabular}
\vspace{-0.1in}
\end{center}
\end{table*}

\begin{table}[h]
\vspace{-0.2in}
\caption{Ablation on the number of clusters $K$ in Eq.~(\ref{eq:training}) on the 2\% subset of QM9. }\label{tab:ablation_k}
\vspace{0.1in}
\begin{center}
\small

\begin{tabular}{lccccc}
\toprule
K   & FCD $\downarrow$   & NSPDK $\downarrow$ & Valid. $\uparrow$ & Unique. $\uparrow$ & Novelty $\uparrow$ \\ 
\midrule
0 & 0.486 & 0.002 & 93.8 & 70.8 & 72.3 \\

1& 0.474 & 0.002& 87.0 & 72.9& 72.0 \\
3&  0.455 & 0.002 & 88.9 & 76.5 & 71.1\\ 
5& {0.443}  & \textbf{0.001} &  88.0 &   {77.0}  &  {73.2}  \\ 
10& {0.434}  & \textbf{0.001} &  \textbf{90.7} &   {75.8}  &  {73.5}  \\ 
20& \textbf{0.430}  & \textbf{0.001} &  87.9 &   \textbf{77.3}  &  {73.8}  \\ 
30& {0.436}  & \textbf{0.001} &  88.9 &   {77.2}  &  \textbf{73.9}  \\ 
{2,113}& {0.443}  & {\textbf{0.001}} &  {86.2} &   {75.4}  &  {72.6}  \\ 
\bottomrule
\end{tabular}

\end{center}

\end{table}

\textbf{Effect of prompt.} In Table~\ref{tab:prompt_ablation}, we show the ablation results on the generation prompt for embedding interpolation-based sampling. We observe that we obtain low FCD and NSPDK scores when we use a prompt similar to the training prompt. However, such choices yield low Novelty scores, generating the many molecules contained in the training samples. The prompt we utilize generates more novel molecules while preserving the state-of-the-art FCD and NSPDK scores. 

\textbf{Effect of hierarchical textual inversion.} In Table~\ref{tab:abla_hier_sup}, we show the importance of our hierarchical textual inversion. Specifically, we compare using a single hierarchy, i.e., detail tokens (using $[D^\ast]$), and our multi-level hierarchy (using $[S^\ast][I^\ast][D^\ast]$). The results show that our multi-level textual inversion strategy is highly useful to generate faithful molecules from the desired distribution.

\textbf{Effect of $K$.} In Table~\ref{tab:ablation_k}, we report the quantitative results of the following cases. First, we consider our proposed design with varying $K$ from 3 to 30. In addition, we consider three other designs that do not contain intermediate tokens to verify the effect of them: (a) $[S_1^\ast][D_n^\ast]$ that the intermediate tokens are removed, i.e., $K$=0, (b) $[S_1^\ast][S_2^\ast][D_n^\ast]$ that the intermediate tokens are replaced with a shared token $[S_2^\ast]$, i.e., $K$=1, {and (c) $[S^\ast][D_{1,n}^\ast][D_{2,n}^\ast]$ that the intermediate tokens are replaced with a detail token $[D_{1,n}^\ast]$, i.e., $K$=2,113.} The results exhibit that the intermediate tokens are indeed crucial for the performance, given that the performance $10\leq K\leq 30$ is much better than (a), (b) and (c). {We find that the overall performance is rather degraded with $K$=2,113 compared to $K$=10, 20, and 30. We hypothesis that this is because the sharing of the coarse-grained common features (i.e., intermediate tokens) serves to regularize the fine-grained features (i.e., detail tokens) which are biased toward a single molecule in the embedding interpolation-based sampling.} We also remark that we did not put much effort on tuning $K$, e.g., $K$=20 improves FCD as $0.434 \rightarrow 0.430$ from $K$=10.

\newpage
\section{Additional Experiments}
\label{sup:additiona_experiments}
\begin{table}[h]
\caption{Generated molecules from \Algname with compositional prompt. We invert 4 aromatic molecules (top row) with the prompt ``The molecule is a $[S^\ast][{D}^\ast_i]$''. With learned embeddings of $[S^\ast]$ and $[{D}^\ast_i]$, we generate molecules (bottom row) with ``The molecule is a boron compound of $[S^\ast][\bar{D}^\ast]$''. We circle the substructures which indicate that the generated molecules indeed satisfy the condition of the given language prompt.}
\vspace{0.1in}
\label{tab:style_transfer}
\begin{center}
\resizebox{0.8\textwidth}{!}{
\scriptsize
\begin{tabular}{cccc}
\toprule

    \multicolumn{4}{c}{Input molecules for inversion}            \\ \midrule
  \includegraphics[height=0.4in,valign=c]{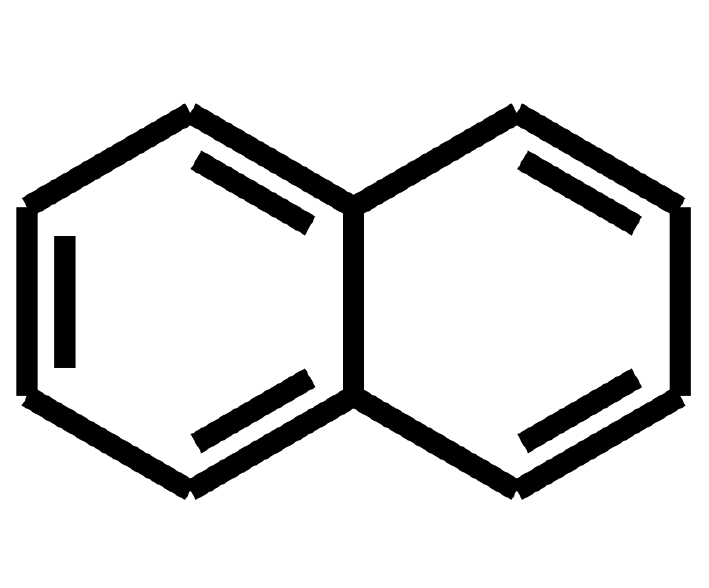}& \includegraphics[height=0.4in,valign=c]{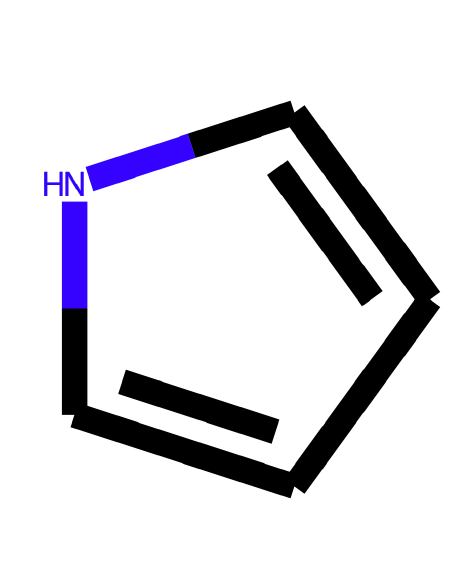}   & \includegraphics[height=0.4in,valign=c]{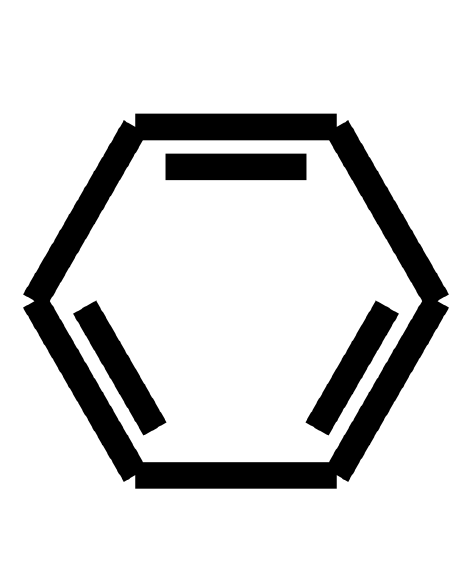} & \includegraphics[height=0.4in,valign=c]{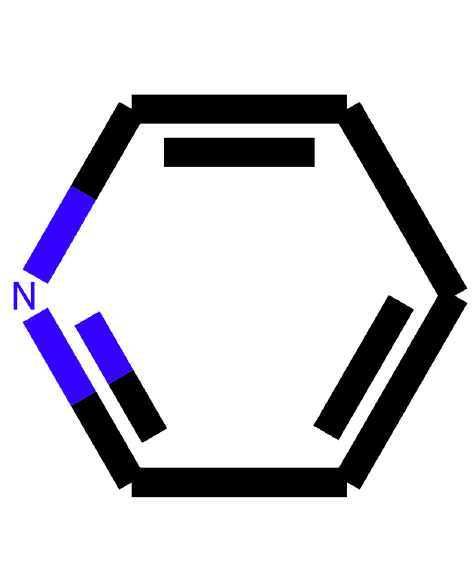} \\
  \multicolumn{4}{c}{The molecule is a $[S^\ast][{D}^\ast_i]$}
  \\\midrule  
\multicolumn{4}{c}{Generated molecules}\\ \midrule

  {\includegraphics[height=0.8in,valign=c]{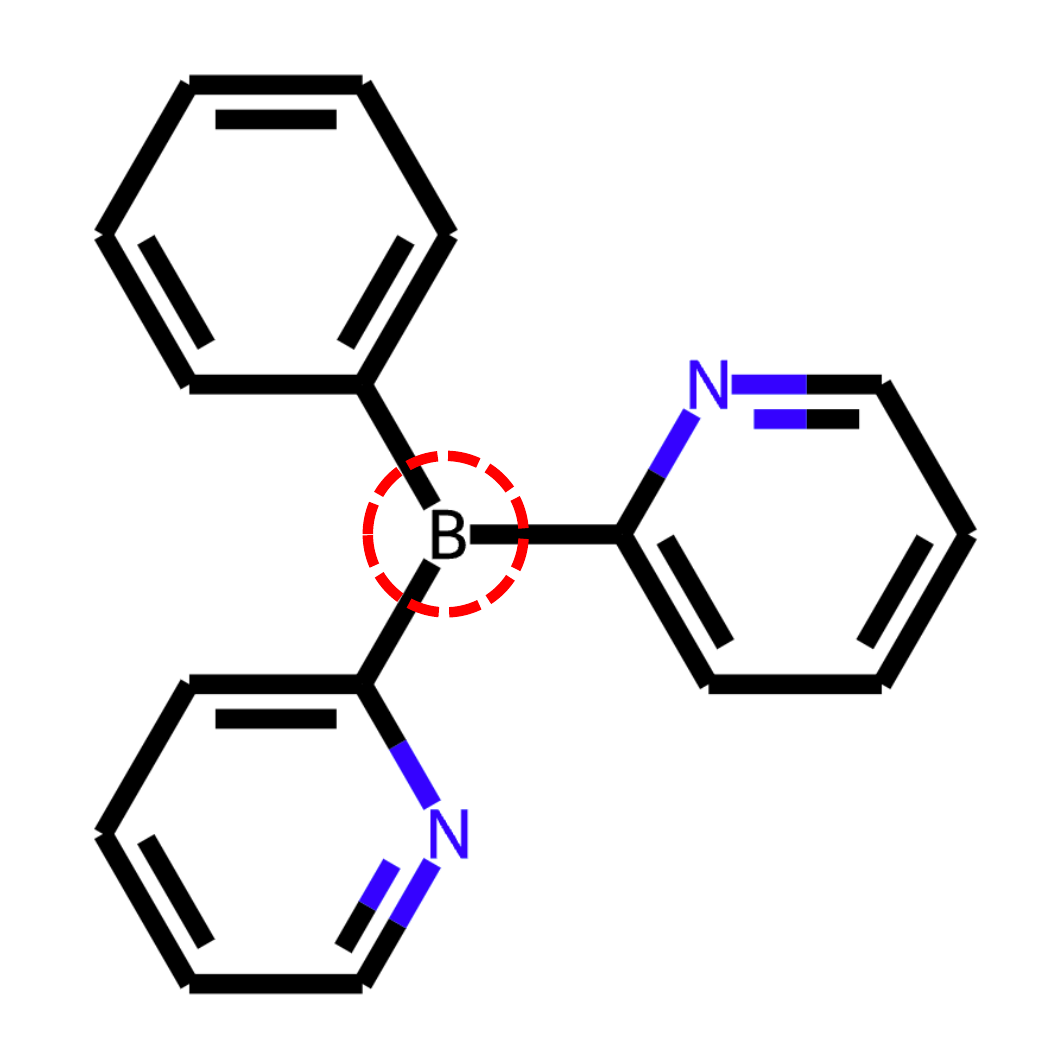}}
  & \includegraphics[height=0.8in,valign=c] {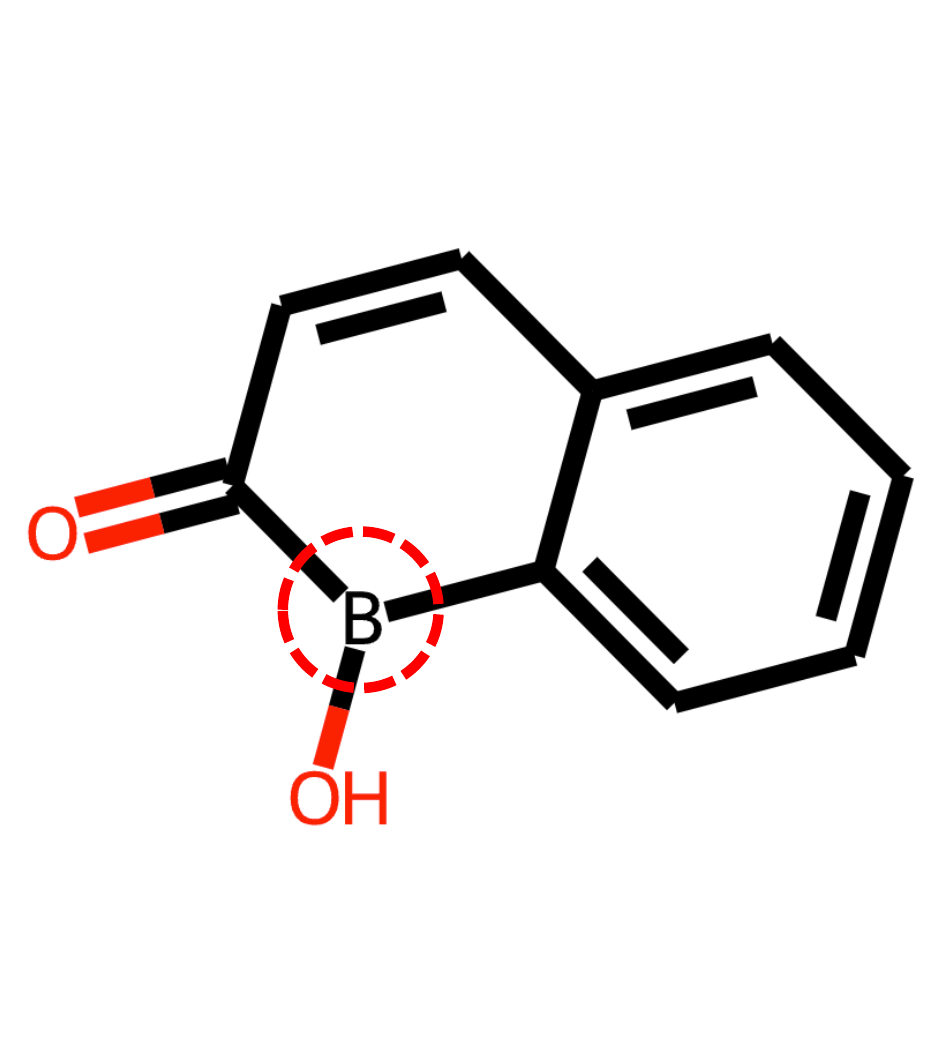}   & \includegraphics[height=0.8in,valign=c]{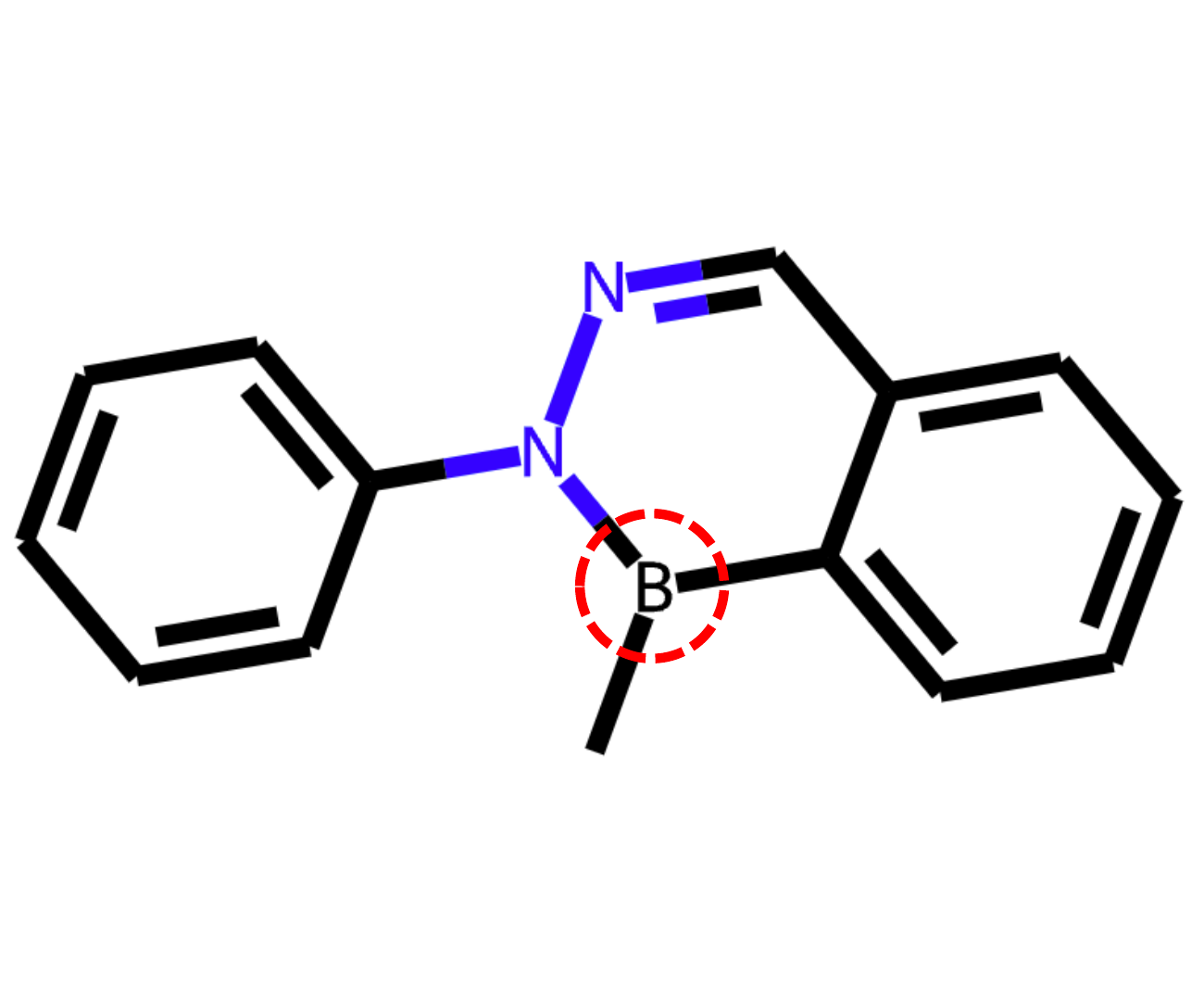} & \includegraphics[height=0.8in,valign=c]{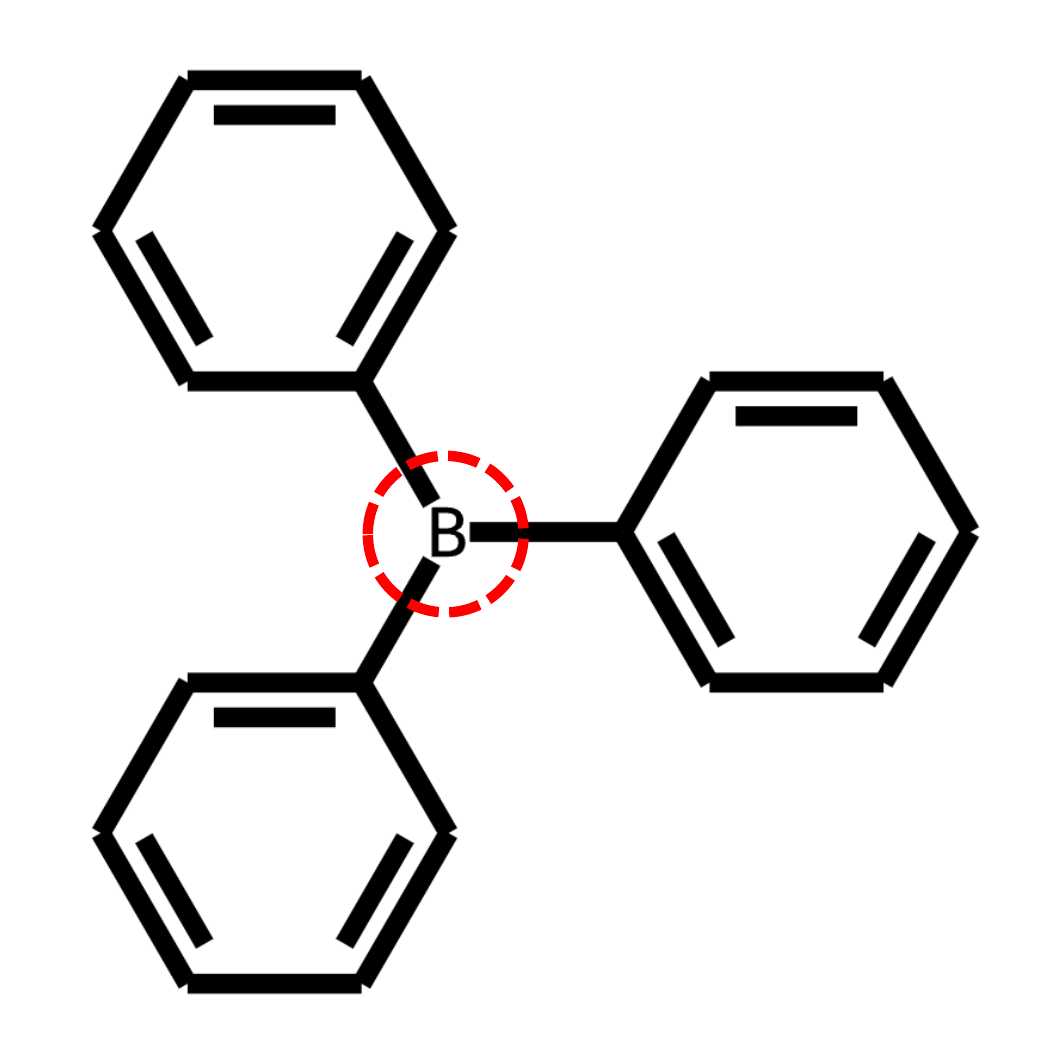} 
  \\
  \multicolumn{4}{c}{The molecule is a \sjline{boron compound} of $[S^\ast][\bar{D}^\ast]$}

\\ \bottomrule
\end{tabular}}
\end{center}
\vspace{-0.1in}
\end{table}

\begin{table}[h]
\vspace{-0.2in}
\caption{Molecular generation results on (1) learning several concepts (the first row) and (2) learning an underlying concept among diverse molecules (the second row).}\label{tab:oracle}

\centering
\scriptsize
\resizebox{0.6\linewidth}{!}{
\begin{tabular}{ccccc}\\\toprule  
&MiCaM  & STGG  & GSDM  &\Algname (Ours)\\\midrule
Success ratio (\%) & 18.2 &33.2 & 0.0 & \textbf{52.0}\\  \midrule
Average QED & 0.555 & 0.558 & 0.090 & \textbf{0.581}\\ 
 \bottomrule
\end{tabular}}
\end{table}

\textbf{Compositionality.} In Table~\ref{tab:style_transfer}, we explore the compositionality of the learned token embeddings from \Algname. We learn the common features of 4 aromatic molecules,\footnote{These molecules share several chemical properties such as resonance and planar structure.} e.g., naphthalene, pyrrole, benzene, and pyridine, via textual inversion. Then, we generate molecules with an additional condition via language prompt. We observe that the generated molecules both satisfy (1) the learned common concept of aromatic molecules and (2) the additional conditions from the language prompt. Although our current molecular language model \citep{edwards2022translation} shows some interesting examples of composition between natural language and the learned concept, we strongly believe that future advances in molecular language models will provide more intriguing examples in this application.

\textbf{Learning complex molecular concepts.} In this section, we explore the ability of \Algname to learn more complex molecular concepts. We conduct two kinds of experiments. Firstly, we impose several target concepts for molecular generation. We collect 300 molecules from GuacaMol \citep{brown2019guacamol} which satisfy QED$>$0.5, SA$>$2.5, and GSK3B$>$0.3.\footnote{QED, SA, and GSK3B measure the drug-likeness, synthesizability, activity to GSK3B, respectively.} With these molecules, we check whether the generative models can learn to model several molecular concepts. We report the ratio of the generated molecules that satisfy the aforementioned condition, e.g., QED$>$0.5, SA$>$2.5, and GSK3B$>$0.3, as the Success ratio in Table~\ref{tab:oracle}. Our \Algname shows superior results on learning several concepts, e.g., $33.2\rightarrow 52.0$, compared to the most competitive baseline, STGG \citep{ahn2022spanning}. Secondly, we explore whether \Algname can learn the ``underlying'' molecular property, e.g., QED, among structurally diverse molecules. We curate 329 molecules in the QM9 dataset \citep{ramakrishnan2014quantum} where (a) each molecule in this subset has a Tanimoto similarity of no higher than 0.4 with any other molecule in the subset and (b) all the molecules in this subset have a high QED ratio greater than 0.6. The average QED in Table~\ref{tab:oracle} shows that \Algname generates molecules with high QED even when the training molecules are structurally largely different, i.e., \Algname indeed learns the underlying molecular concept.

\newpage

\begin{table}[h]
\caption{Comparison with pre-trained model of STGG \citep{ahn2022spanning} on the HIV dataset.}
\label{tab:moleculenet_pretrained}
\begin{center}
\vspace{0.1in}
\small%
\begin{tabular}{lcccccc}
\toprule
     Method &   Active. $\uparrow$ & FCD $\downarrow$ & NSPDK $\downarrow$ & Valid. $\uparrow$ & Unique. $\uparrow$ & Novelty $\uparrow$ \\ \midrule

STGG (from scratch) & 1.6 & 20.2 & 0.033 & \textbf{100} & \textbf{95.8} & \textbf{100}   \\ 
STGG (fine-tuned) & 3.6 & 20.0 & 0.030 & \textbf{100} & 87.1  & \textbf{100} \\
 \midrule
\cellcolor{tablegreen}\textbf{\Algname (Ours)} &  \cellcolor{tablegreen}\textbf{11.4}  &\cellcolor{tablegreen} \textbf{16.6} & \cellcolor{tablegreen}\textbf{0.019}&\cellcolor{tablegreen}\textbf{100} &\cellcolor{tablegreen}95.6&\cellcolor{tablegreen}\textbf{100}\\

\bottomrule

\end{tabular}

\end{center}

\end{table}

\textbf{Comparison with pre-trained model.} In Table~\ref{tab:moleculenet_pretrained}, we report the performance of the baseline method by fine-tuning the pre-trained baseline model. Specifically, we fine-tune the model of STGG \citep{ahn2022spanning} pre-trained with the ZINC250k dataset \citep{irwin2012zinc} on the HIV dataset \citep{wu2018moleculenet}. We observe that \Algname still achieves significantly better performance in overall metrics, e.g., 20.0 $\rightarrow$ 16.6 and 0.030 $\rightarrow$ 0.019 in FCD and NSPDK, respectively.

\section{Modification Algorithm}
\label{sup:modification}
\newcommand\mycommfont[1]{\footnotesize\ttfamily\textcolor{blue}{#1}}
\SetCommentSty{mycommfont}

\SetKwInput{KwInput}{Input}                
\SetKwInput{KwOutput}{Output}              
\SetKwData{Exist}{exist}
\SetKwData{Bct}{branch closing token}
\SetKwData{Bot}{branch opening token}
\begin{algorithm}[!ht]
\DontPrintSemicolon
  
  \KwInput{An invalid SMILES string}
  \KwOutput{A modified SMILES string}
  
  \While{exist a branch closing token token prior to a branch opening token}
  {
    Remove the corresponding branch closing token. \tcp*{``CC)CCC'' to ``CCCCC''}
  }
  \While{exist an unclosed branch opening token}
  {
    Add the the branch closing token at the end of the string. \tcp*{``CC(CCC'' to ``CC(CCC)''}
  }
  \While{exist an unclosed ring opening token}
  {
    Remove the ring opening token. \tcp*{``CC1CCC'' to ``CCCCC''}
  }
  \While{exist an atom that exceeds the valency}
  {
    Randomly drop a branch to satisfy the valency. \tcp*{``C\#C(=CC)C to ``C\#CC''}
  }
  \While{exist a ring with less than 3 atoms}
  {
    Remove the ring opening/closing token. \tcp*{``CC1C1 to ``CCC''}
  }
  
\caption{Modification algorithm for an invalid SMILES string}
\end{algorithm}

\newpage
\section{Details on QM9 Experiments}
\label{appen:qm9}

\begin{table}[h]

\caption{Qualitative results for molecular generation varying the data ratio on QM9.}\label{tab:ratio_ablation}
\vspace{0.1in}
\begin{center}
\small

\begin{tabular}{llcccccc}
\toprule
Ratio $(\%)$ & Method & Grammar  & FCD $\downarrow$   & NSPDK $\downarrow$ & Valid. $\uparrow$ & Unique. $\uparrow$ & Novelty $\uparrow$ \\ 
\midrule
\multirow{4.5}{*}{2}& GDSS \citep{jo2022score} & \textcolor{SJRed}{\xmark} & 22.953 & 0.455 & 99.8 & {1.2} & {72.2} \\
& STGG \citep{ahn2022spanning} & \textcolor{SJViolet}{\cmark} & 0.715 & 0.002 & \textbf{100} & \textbf{88.0} & {42.1} \\\cmidrule{2-8}
& \textbf{\Algname (Ours)} & \textcolor{SJRed}{\xmark} & 0.434 & \textbf{0.001}& {90.7} & 75.8& {73.5} \\
             & \textbf{\Algname (Ours)}     &   \textcolor{SJViolet}{\cmark}    & {0.430}  & \textbf{0.001} &  \textbf{100} &   76.1  &  \textbf{75.6}  \\  \midrule
\multirow{4.5}{*}{5}& GDSS \citep{jo2022score} & \textcolor{SJRed}{\xmark} & 17.013 & 0.066 & 97.2 & 25.5 & {44.2} \\
& STGG \citep{ahn2022spanning} & \textcolor{SJViolet}{\cmark} & 0.665 & \textbf{0.001} & \textbf{100} & \textbf{95.8} & 63.0\\ \cmidrule{2-8}
& \textbf{\Algname (Ours)} & \textcolor{SJRed}{\xmark} & 0.412 & \textbf{0.001} & 89.4 & 85.8 & 70.4\\ 
                  & \textbf{\Algname (Ours)}&       \textcolor{SJViolet}{\cmark}    & {0.410}  & \textbf{0.001} & \textbf{100} & 86.4  &  \textbf{72.4}  \\  \midrule
\multirow{4.5}{*}{10}& GDSS \citep{jo2022score} & \textcolor{SJRed}{\xmark} & 17.170 & 0.067 & 98.0 & 22.8 & 36.6 \\
& STGG \citep{ahn2022spanning} &\textcolor{SJViolet}{\cmark} & 0.603 & 0.002 & \textbf{100} & \textbf{99.4} & {63.5}  \\ \cmidrule{2-8}
&\textbf{\Algname (Ours)}  &\textcolor{SJRed}{\xmark} & {0.400}  & {0.002} &  87.6 &   {87.6}  &  {71.2}  \\ 
                  &\textbf{\Algname (Ours)}&           \textcolor{SJViolet}{\cmark}& {0.398}  & \textbf{0.001} &  \textbf{100} &   {88.3}  &  \textbf{73.2}  \\ \midrule
\multirow{4.5}{*}{20}& GDSS \citep{jo2022score} & \textcolor{SJRed}{\xmark} & 7.345 & 0.025 & 94.2 & 82.3 & 67.6 \\
& STGG \citep{ahn2022spanning} & \textcolor{SJViolet}{\cmark} & 0.599 & \textbf{0.001} & \textbf{100} & \textbf{99.4} &{64.3}  \\ \cmidrule{2-8}
& \textbf{\Algname (Ours)} & \textcolor{SJRed}{\xmark} & {0.384}  & \textbf{0.001} &  86.7 &   {87.8}  &  {70.0}  \\ 
                   & \textbf{\Algname (Ours)}&          \textcolor{SJViolet}{\cmark}& {0.383}  & \textbf{0.001} &  \textbf{100} &   {88.7}  & \textbf{71.8}  \\\midrule
\multirow{4.5}{*}{50}& GDSS \citep{jo2022score} & \textcolor{SJRed}{\xmark} &3.564& 0.008 & 96.0 & 96.6 & 80.1 \\
& STGG \citep{ahn2022spanning} & \textcolor{SJViolet}{\cmark} & 0.592 & \textbf{0.001} & \textbf{100} & \textbf{99.2} & \textbf{70.6}\\ \cmidrule{2-8}
& \textbf{\Algname (Ours)}  & \textcolor{SJRed}{\xmark} & \textbf{0.372} & \textbf{0.001} & 88.7 & 87.7 & 68.8\\
                   &  \textbf{\Algname (Ours)} &       \textcolor{SJViolet}{\cmark}& \textbf{0.372} & \textbf{0.001} & \textbf{100} & 88.5 & 70.5 \\

\bottomrule
\end{tabular}
\end{center}

\vspace{-0.3in}
    \end{table}

\begin{table}[h]
\caption{Comparison with the baseline with high Novelty via resampling strategy on QM9.}
\vspace{0.1in}
\label{tab:qm9_resampling}
\begin{center}
\small

\begin{tabular}{lcccccc}
\toprule
     Method & Resampling ratio &  FCD $\downarrow$ & NSPDK $\downarrow$ & Valid. $\uparrow$ & Unique. $\uparrow$ & Novelty $\uparrow$ \\ \midrule

GDSS \citep{jo2022score} & 1.0 & 2.900 & 0.003 & {95.7} & {98.5} & {86.3}   \\ 
 \midrule
\cellcolor{tablegreen}\textbf{\Algname (Ours; 2\%)} &  \cellcolor{tablegreen}{1.9}  &\cellcolor{tablegreen} \textbf{0.601} &  \cellcolor{tablegreen}\textbf{0.002}& \cellcolor{tablegreen}\textbf{100} & \cellcolor{tablegreen}\textbf{100}& \cellcolor{tablegreen}\textbf{100}\\

\bottomrule

\end{tabular}

\end{center}

\end{table}

In Table~\ref{tab:ratio_ablation}, we report experimental results varying the data ratio from 2\% to 50\%. In particular, when we use 50\% of the training data the performance improves further by 0.430 $\rightarrow$ 0.372 (compared to using 2\% of training data), i.e., our \Algname better learns molecule distribution when more molecules are available for training.

We note that there is a fundamental trade-off between FCD and Novelty. If the generated molecules have many overlaps with training molecules, i.e., low Novelty, the FCD score improves, i.e., decreases, since the generated molecules are more likely to follow the target distribution. Therefore, it is crucial to compare FCD under a similar Novelty score. Therefore, in Table~\ref{tab:qm9_resampling}, we report the generation results with the resampling strategy, i.e., we sample molecules until we have 10,000 molecules with Validity, Uniqueness, and Novelty scores as 100 and we reject samples that violate these scores. We denote the relative ratio of the total sampling trial (including the rejected ones) as Resampling ratio. Here, we remark that such resampling process does not incur much computational cost, e.g., only 1.8 sec for a sample (see Appendix~\ref{sup:complexity} for analysis of time complexity). The result shows that \Algname generates high-quality novel molecules from our desired target distribution.

\newpage

\section{Analysis of Interpolation-based Sampling}
\label{appen:interpolation_analysis}
\begin{table}[h]
\caption{Generated molecules from \Algname with varying $\lambda$ in Eq. (\ref{eq:lambda}). Samples are generated with the prompt ``A similar chemical of $[S^\ast][\bar{I}^\ast][\bar{D}^\ast]$''. The columns $[{D}^\ast_i]$ and $[{D}^\ast_j]$ denote molecules in the HIV dataset \citep{wu2018moleculenet} whose token embeddings are interpolated for each row.}
\label{tab:lbd_result}
\vspace{0.1in}
\resizebox{1.0\textwidth}{!}{
\LARGE 
\begin{tabular}{c|ccccc|c}
\toprule

    {$[D^\ast_i]$}           & \multicolumn{5}{c|}{A similar chemical of $[S^\ast][\bar{I}^\ast][\bar{D}^\ast]$ }                      & $[D^\ast_j]$    \\ \midrule
  \includegraphics[height=0.8in,valign=c]{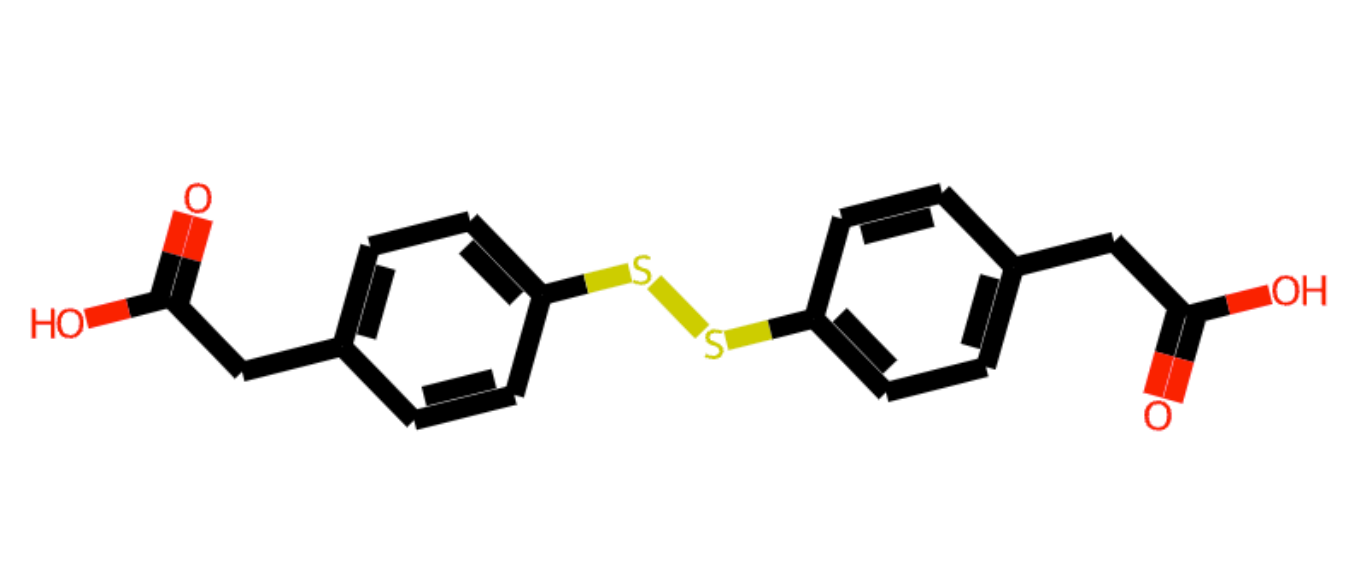}& \includegraphics[height=0.8in,valign=c]{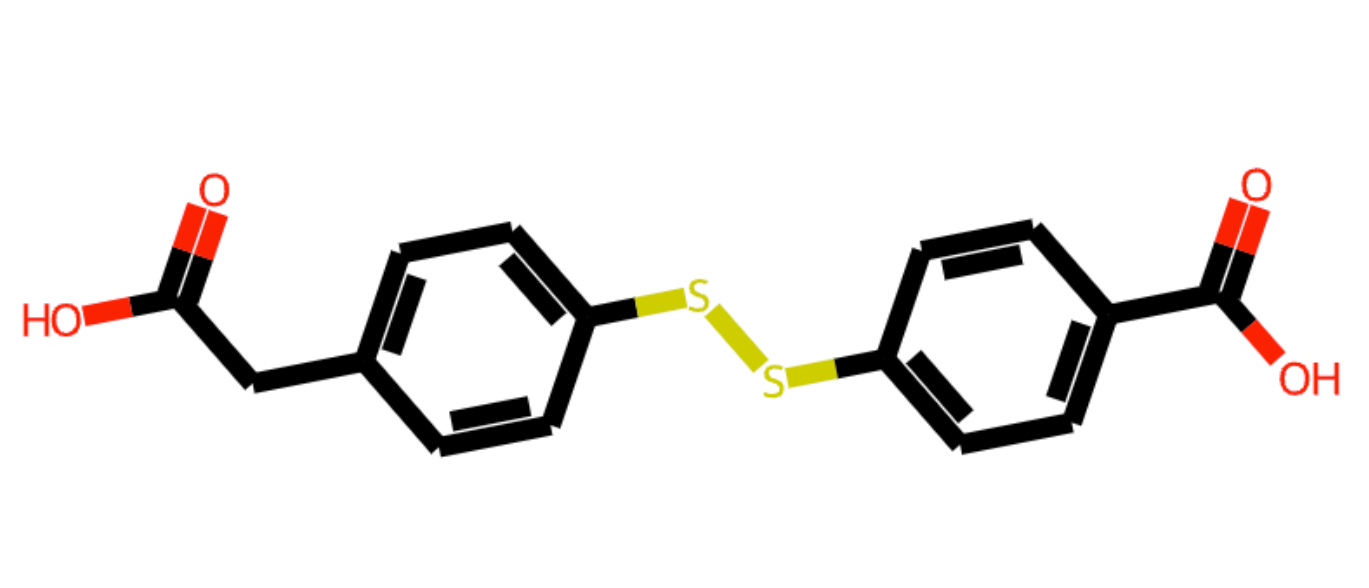}   & \includegraphics[height=0.8in,valign=c]{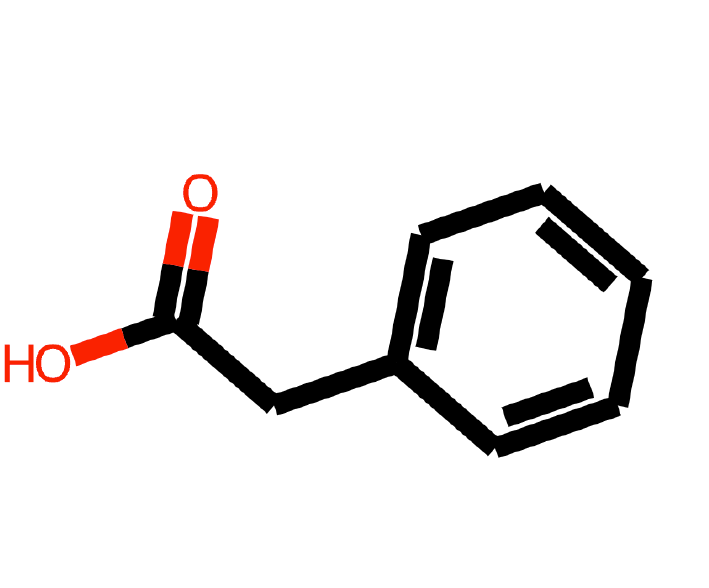} & \includegraphics[height=0.8in,valign=c]{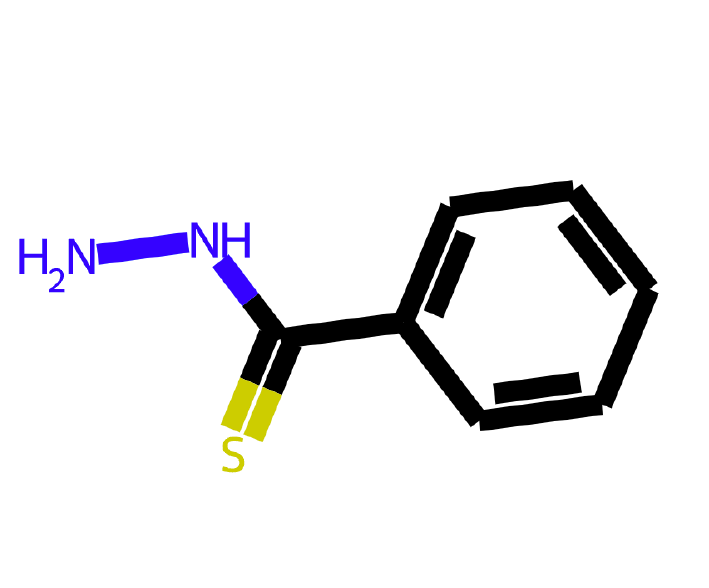} & \includegraphics[height=0.8in,valign=c]{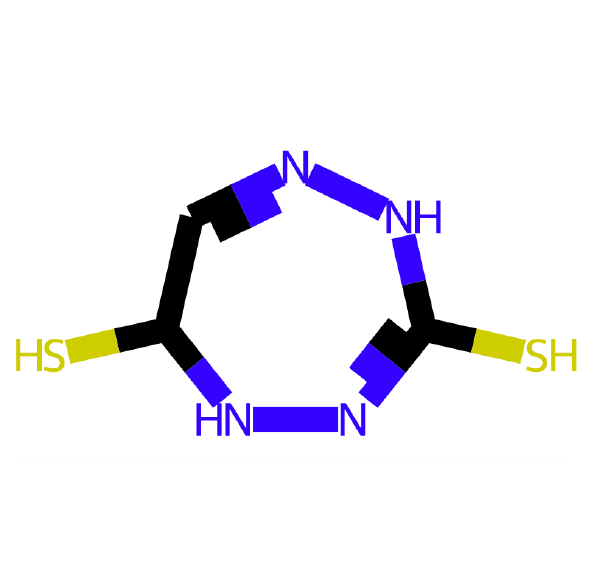} &
  \includegraphics[height=0.8in,valign=c]{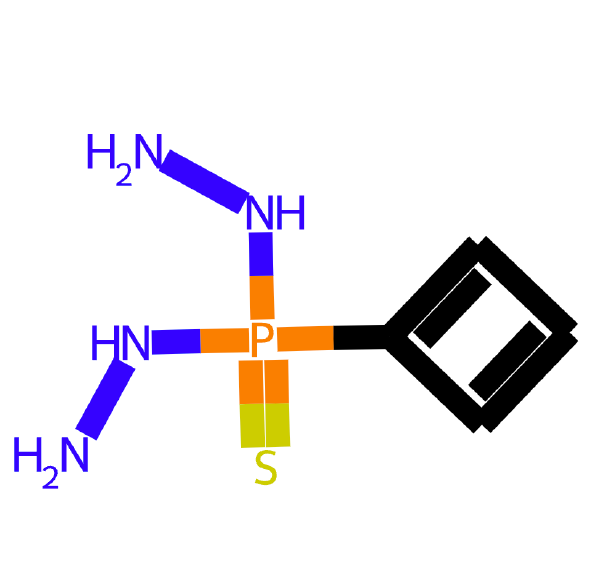}     &  \includegraphics[height=0.8in,valign=c]{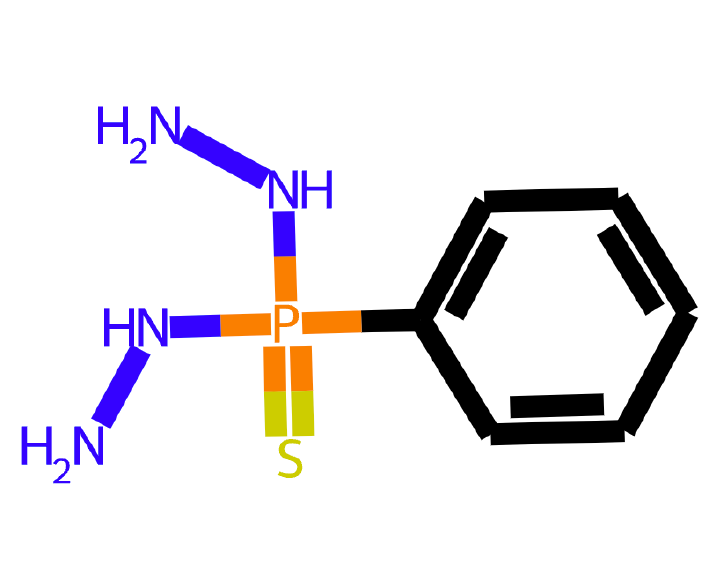}\\ 
  &  $\lambda=0.0$   & {\phantom{0000}$\lambda=0.3$\phantom{0000}}  & {\phantom{0000}$\lambda=0.5$\phantom{0000}} & {\phantom{00000}$\lambda=0.7$\phantom{00000}}  & {\phantom{00000}$\lambda=1.0$\phantom{0000}}    \\ \midrule  

  {\includegraphics[height=0.8in,valign=c]{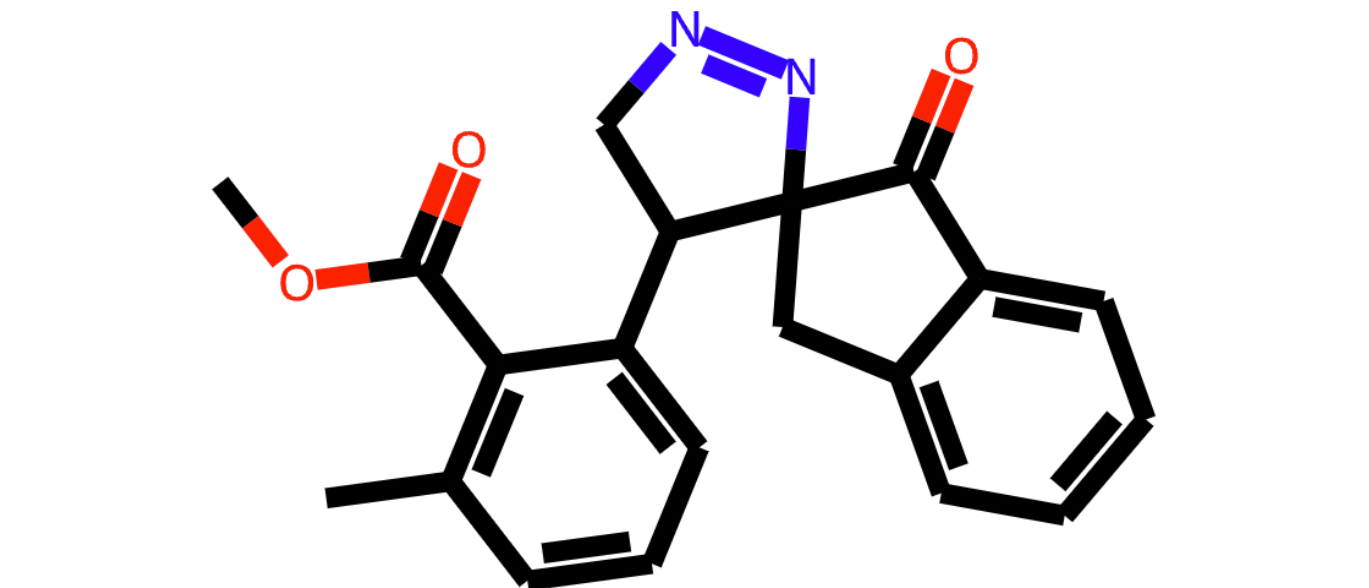}}& \includegraphics[height=0.8in,valign=c]{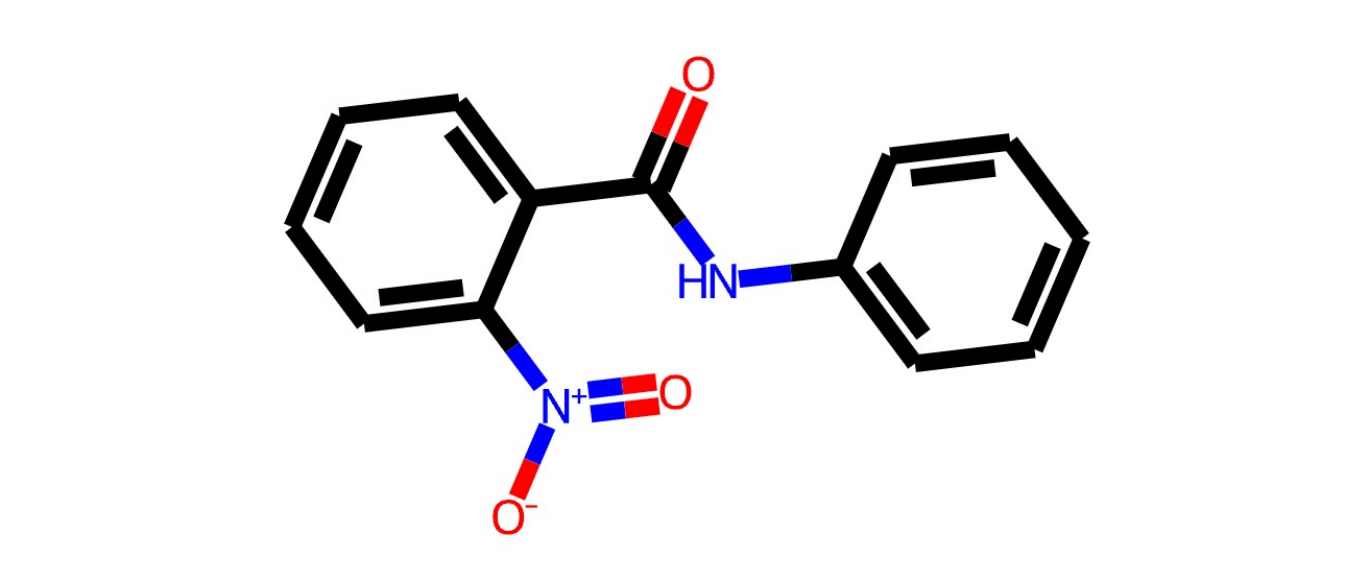}   & \includegraphics[height=0.8in,valign=c]{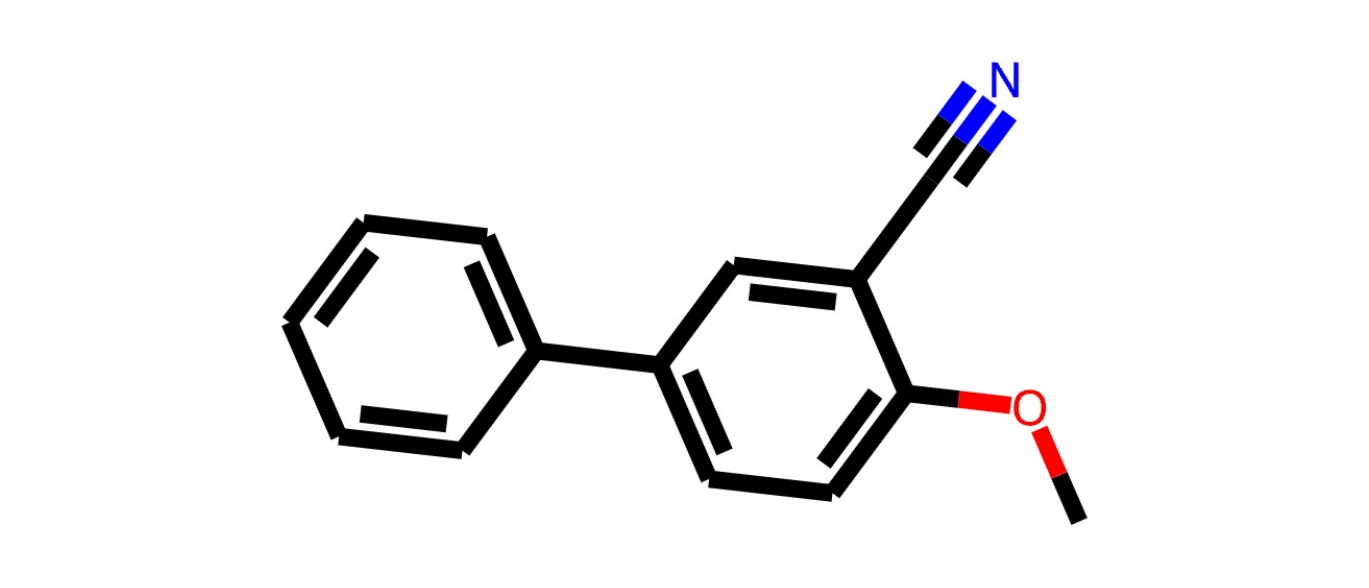} & \includegraphics[height=0.8in,valign=c]{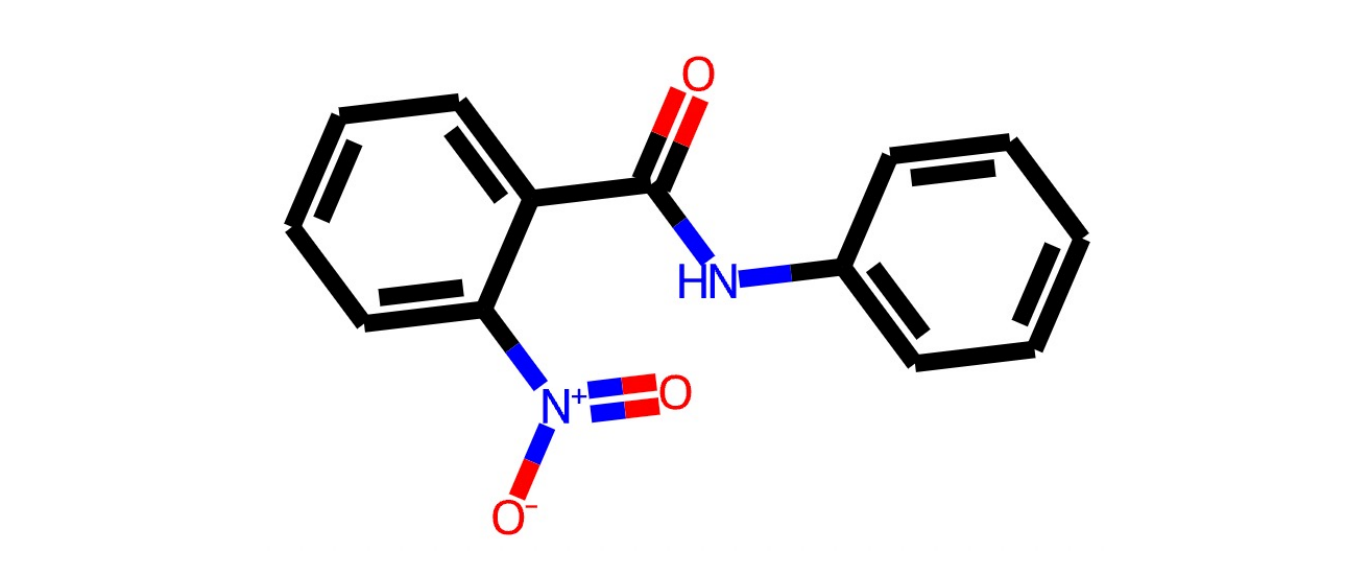} & \includegraphics[height=0.8in,valign=c]{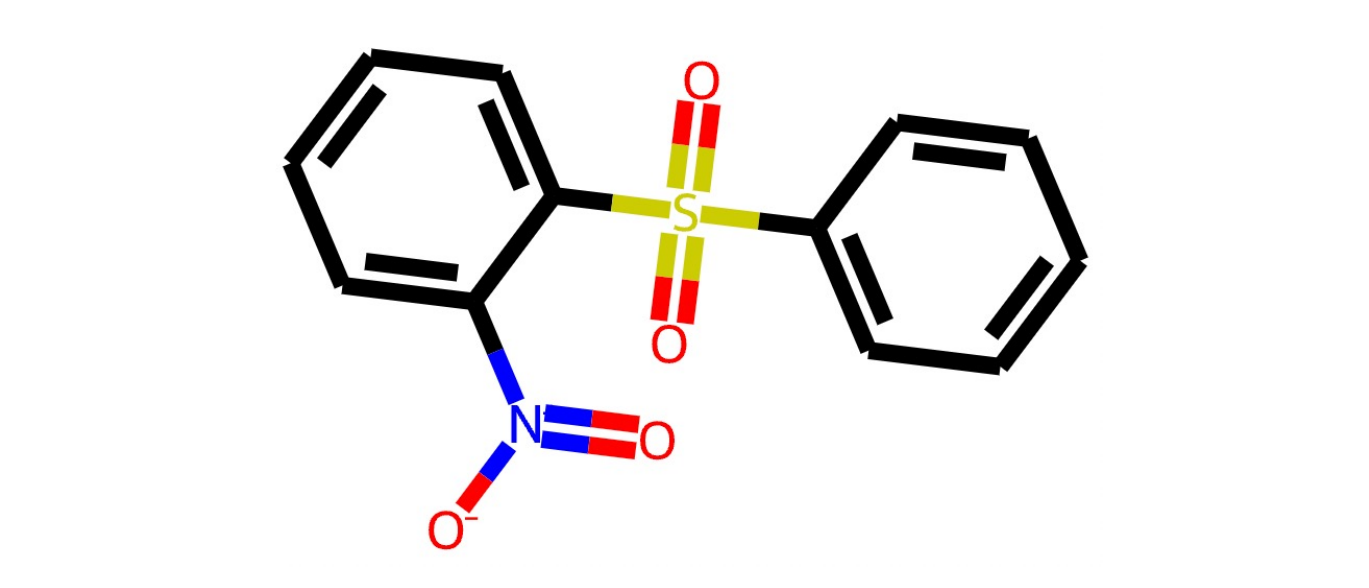} &
  \includegraphics[height=0.8in,valign=c]{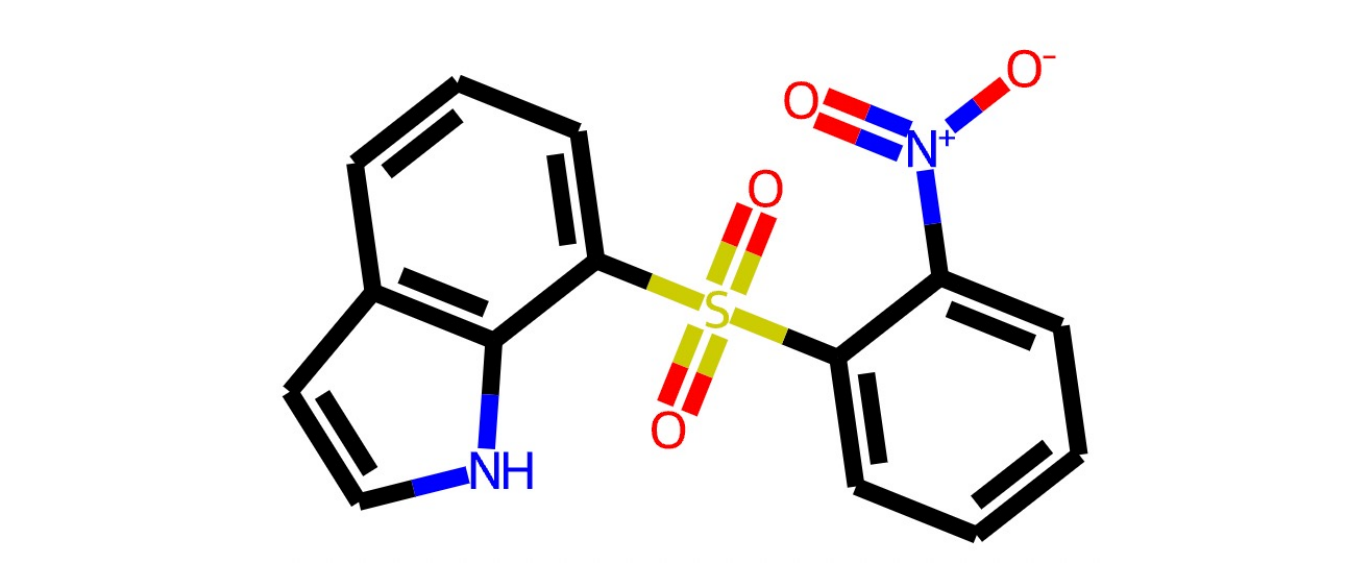}     &  \includegraphics[height=0.8in,valign=c]{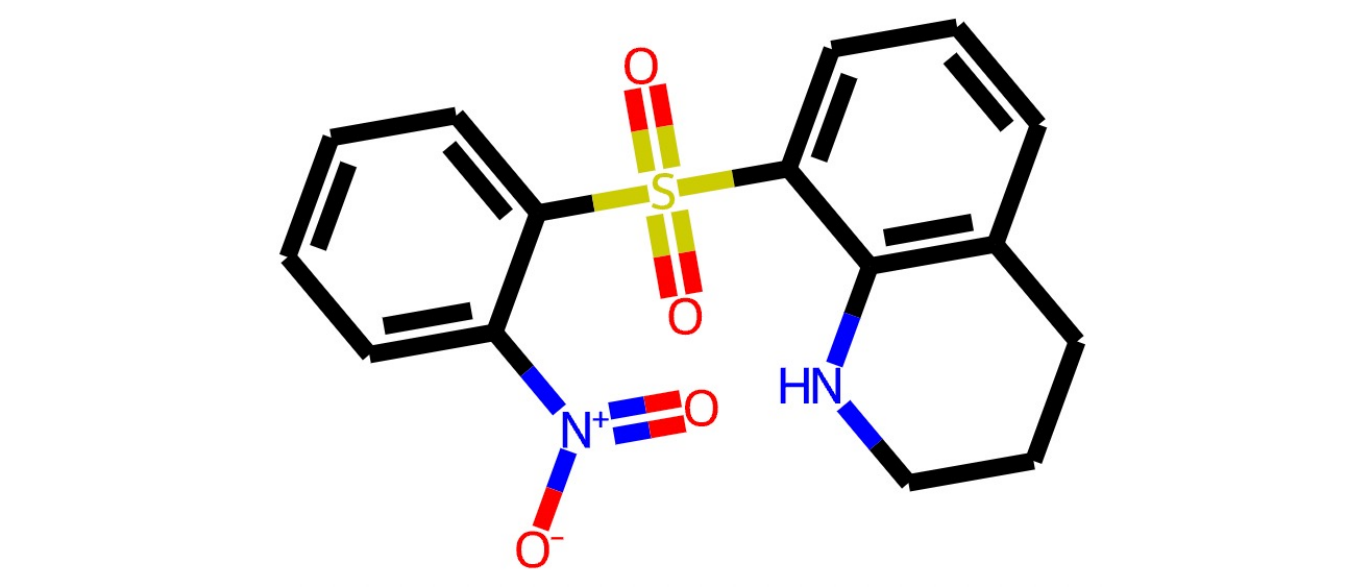}\\ 
  &  $\lambda=0.0$   & {\phantom{0000}$\lambda=0.3$\phantom{0000}}  & {\phantom{0000}$\lambda=0.5$\phantom{0000}} & {\phantom{00000}$\lambda=0.7$\phantom{00000}}  & {\phantom{00000}$\lambda=1.0$\phantom{0000}}      
\\ \bottomrule
\end{tabular}}
\vspace{-0.2in}
\end{table}
\begin{table}[h]
\caption{Generated molecules from \Algname with varying $\lambda$ in Eq. (\ref{eq:lambda}). We interpolate a single-level token, e.g., ``A similar chemical of $[S^\ast][\bar{I}^\ast][{D}^\ast]$'' and ``A similar chemical of $[S^\ast][{I}^\ast][\bar{D}^\ast]$''.}
\label{tab:lbd_single_token}
\vspace{0.1in}
\centering
\resizebox{0.7\textwidth}{!}{ 
\begin{tabular}{c}
\toprule

    {A similar chemical of $[S^\ast][\bar{I}^\ast][{D}^\ast]$ }                \\ \midrule
   \includegraphics[height=0.8in,valign=c]{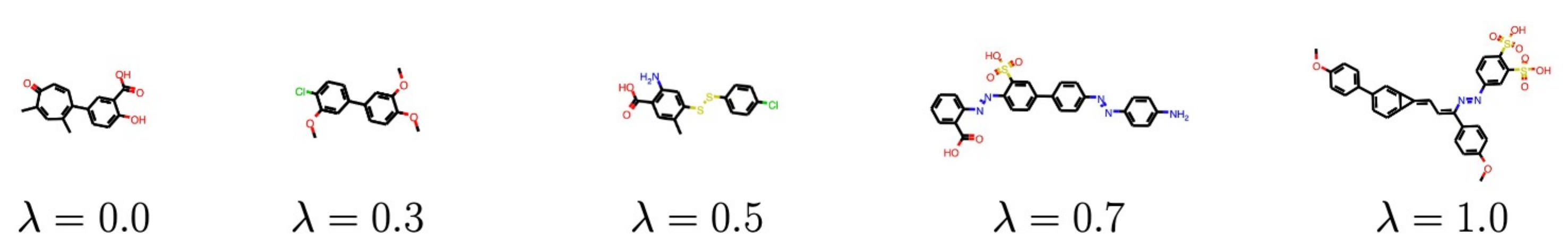}\\ 
  \midrule  
      {A similar chemical of $[S^\ast][{I}^\ast][\bar{D}^\ast]$ }                \\ \midrule
      \includegraphics[height=0.8in,valign=c]{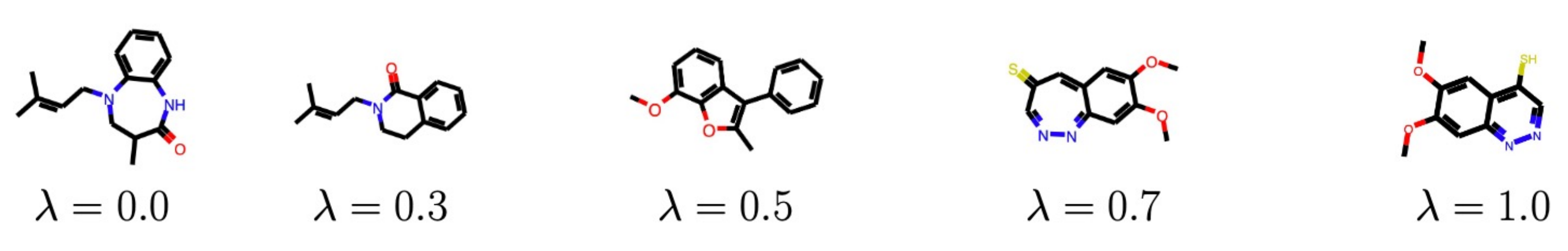}
      
\\ \bottomrule
\end{tabular}
}

\end{table}
Note that our sampling is based on the interpolation of two different token embeddings with different values of $\lambda \sim p(\lambda)$. In Table~\ref{tab:lbd_result}, we provide how the generated molecules are changed with different values of $\lambda$. With varying $\lambda$, one can observe that the generated molecules (1) maintain some original important low-level semantics and (2) introduce some novel aspects distinct from both original semantics. For example, $\lambda=0.7$ in the first row of Table~\ref{tab:lbd_result} introduces a new 4-membered ring system while preserving the phosphorous-sulfur double bond structure of the original features in $[D^\ast_j]$. This observation exhibits that our embedding space models the manifold of underlying target distribution effectively, enabling data-efficient sampling from the target distribution. We also provide the generated samples from different hierarchies. Interpolating intermediate tokens (see the first row of Table~\ref{tab:lbd_single_token}) change the low-level semantics, i.e., size of molecules, of the generated molecules and interpolating detail (see the second row of Table~\ref{tab:lbd_single_token}) tokens change the high-level features, i.e., insertion of a single atom, of the generated molecules.

\newpage
\section{Complexity}
\label{sup:complexity}

\begin{table}[h]
\caption{Time and space complexity of each molecular generative method.}
\centering\small
\vspace{0.1in}
\label{tab:time_complexity}
\resizebox{1.0\textwidth}{!}{%
\begin{tabular}{lccccccccc}
\toprule
    & JT-VAE & PS-VAE & MiCaM & STGG & CRNN & GDSS & GSDM & DiGress & HI-Mol (Ours) \\ 
\midrule
Time complexity (s)&4.8 & 0.1 & 0.9 &0.7 & 0.5 & 71.2 & 2.0 & 9.1 & 1.8\\
Space complexity (GB) & 0.4 & 1.2 & 1.6 & 2.1 & 0.4 & 1.2 &1.1 & 1.5 & 4.8\\
\bottomrule
\end{tabular}
}
\vspace{-0.05in}
\end{table}
In Table~\ref{tab:time_complexity}, we provide the time and space complexity to generate a molecule via various molecular generative models. For time complexity, measured with a single RTX 3090 GPU, HI-Mol takes about 1.8 seconds to sample a single molecule, while other methods, e.g., GDSS and DiGress, require more time due to denoising diffusion steps. For memory complexity, HI-Mol requires 4.8GB of GPU VRAM space due to the usage of the large model. We believe that reducing this space for large language models, e.g., through \citet{dao2022flashattention}, will be an interesting future direction.

\section{Discussion on Molecular Optimization}
\label{sup:optimization}

\begin{wrapfigure}{r}{0.35\textwidth}
\vspace{-0.15in}
\includegraphics[width=\linewidth]{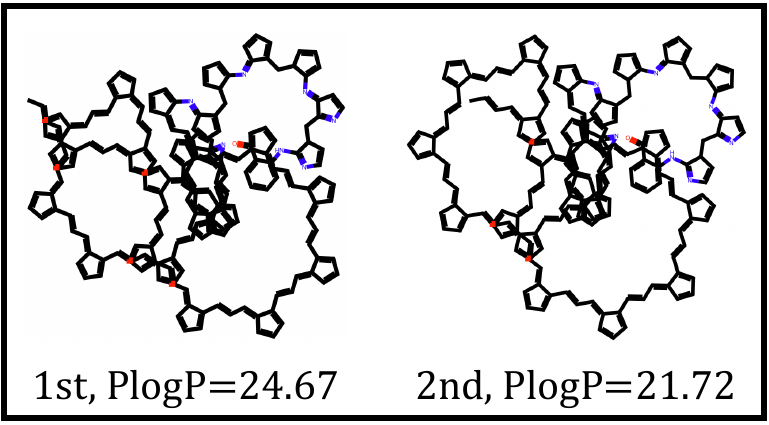} 
\caption{Visualizations of the generated molecules with $\gamma=50$. The maximum PLogP among the training molecules is 4.52.}
\label{fig:optimization_unrealistic}
\vspace{-0.1in}
\end{wrapfigure}

In Table~\ref{tab:optimization}, we have shown the usefulness of our \Algname to maximize the PLogP value of the generated molecules. While this evaluation setup for molecular optimization is a common and popular choice in molecular domain \citep{jin2018junction,shi2020graphaf,luo2021graphdf,ahn2022spanning}, some prior works have noted that solely maximizing the PLogP value may yield unstable or hard-to-synthesize molecules \citep{gao2020synthesizability,coley2021defining,ahn2022spanning}. In Figure~\ref{fig:optimization_unrealistic}, we show the visualizations of the optimized molecules with the highest PLogP values. Similar to the most competitive baseline, STGG \citep{ahn2022spanning}, our optimized molecules contain a large number of atoms, and thus relatively hard to synthesize. Although these results show that our \Algname effectively learns to incorporate the condition PLogP in a data-efficient manner, it would be an important research direction to develop an evaluation framework for molecular optimization that takes into account the ``realistic-ness'', e.g., stability and synthesizability, of the molecules.

\newpage
\section{Details on Low-shot Molecular Property Prediction}
\label{appen:classification_detail}

\begin{table}[h]
\caption{Results on low-shot classification on the MoleculeNet benchmark. We report the average and 95\% confidence interval of the test ROC-AUC scores within 20 random seeds.}\label{tab:classification_confidence}
\vspace{0.1in}
\begin{center}
\small

\begin{tabular}{clcc}
\toprule
{Dataset}                      & \multicolumn{1}{c}{Method}          & 16-shot                & 32-shot           
\\ \midrule
\multirow{4}{*}{HIV}           
                               & DiGress \citep{vignac2023digress}   & -2.30\stdv{3.50}      & -2.67\stdv{3.15}  \\
                               & MiCaM \citep{geng2023novo}     & 1.02\stdv{3.29}       & 0.69\stdv{2.09}   \\
                               
                               & STGG \citep{ahn2022spanning}        & 0.53\stdv{2.79}       & -0.47\stdv{2.36}  \\
                               \cmidrule{2-4} 
                               & \cellcolor{tablegreen}\textbf{HI-Mol (Ours)}              & \cellcolor{tablegreen}\textbf{2.35}\stdv{2.71}  & \cellcolor{tablegreen}\textbf{2.16}\stdv{1.64} \\ 
                               \midrule
\multirow{4}{*}{BBBP}          
                               & DiGress \citep{vignac2023digress}   & 1.73\stdv{1.53}          & 0.97\stdv{1.99} \\
                               & MiCaM \citep{geng2023novo}     & 1.91\stdv{2.13}         & 1.78\stdv{1.98} \\
                               
                               & STGG \citep{ahn2022spanning}        & 1.85\stdv{1.83}          & 1.76\stdv{1.72} \\
                               \cmidrule{2-4} 
                               & \cellcolor{tablegreen}\textbf{HI-Mol (Ours)}              & \cellcolor{tablegreen}\textbf{2.73}\stdv{2.01}  & \cellcolor{tablegreen}\textbf{2.64}\stdv{1.75} \\ \midrule
\multirow{4}{*}{BACE}          
                               & DiGress \citep{vignac2023digress}   & -0.60\stdv{2.88}       & -0.91\stdv{1.82}  \\
                               & MiCaM \citep{geng2023novo}     & -0.65\stdv{3.17}       & -1.11\stdv{2.95}  \\
                               
                               & STGG \citep{ahn2022spanning}        & 2.34\stdv{2.15}        & 2.01\stdv{1.45}    \\
                               \cmidrule{2-4} 
                               & \cellcolor{tablegreen}\textbf{HI-Mol (Ours)}              & \cellcolor{tablegreen}\textbf{3.53}\stdv{1.57}  & \cellcolor{tablegreen}\textbf{3.39}\stdv{1.80}  \\ \bottomrule

\end{tabular}

\end{center}
\end{table}

\begin{table}[h]
\caption{Comparison with latent mixup \citep{wang2021mixup} in the low-shot classification task. We report $\Delta$ROC-AUC averaged over 20 random seeds.}\label{tab:classification_mixup}
\vspace{0.1in}
\begin{center}
\small

\begin{tabular}{lccc}
\toprule
32-Shot         & HIV          & BBBP                & BACE           
\\ \midrule
{Latent mixup \citep{wang2021mixup}}           
                               & 0.55   & 1.27      & 0.52  \\
                               \midrule
\cellcolor{tablegreen}{\textbf{\Algname (Ours)}}          
                               & \cellcolor{tablegreen}\textbf{2.16}   & \cellcolor{tablegreen}\textbf{2.64}          & \cellcolor{tablegreen}\textbf{3.39} \\
                                 \midrule

\end{tabular}

\end{center}
\end{table}

Low-shot (or few-shot) prediction tasks are one of the important applications for industrial deployments \cite{nam2023stunt}, and we have shown our \Algname's capability to be beneficial to these tasks. 
In Table~\ref{tab:classification_confidence}, we report the full results of low-shot molecular property prediction experiments with averages and 95\% confidence intervals.
With randomly sampled low-shot molecules from the train split (used in our main experiments of Table~\ref{tab:moleculenet}), we generate $\times$3 number of valid molecules via generative models, e.g., we generate 96 molecules for 32-shot experiments.
For the classifier, we utilize the 5-layer GIN \citep{xu2018how} from \citet{you2020graph}, which is pre-trained with unlabeled molecules via self-supervised contrastive learning. We fine-tune this model for 100 epochs by introducing a linear projection head for each dataset. We use Adam optimizer with a learning rate of 0.0001 and no weight decay. The results are calculated based on the test ROC-AUC score of the epoch with the best validation ROC-AUC score. Specifically, we consider two scenarios: (1) training the classifier with only the low-shot molecules and (2) training the classifier with both the original low-shot molecules and the generated molecules via the molecular generative model. We report $\Delta$ROC-AUC score, calculated by the subtraction of the ROC-AUC score of (1) from (2). In Table~\ref{tab:classification_mixup}, we additionally compare with conventional latent mixup strategy \citep{wang2021mixup}. They directly use the interpolated latent embeddings (and corresponding interpolated labels) as inputs, which mostly do not become real data. However, we generate ``new molecules'' (rather than just latent embeddings) based on this embedding and use it as real data to train a classifier for a molecular prediction task. For the latent mixup, we train the classifier using given molecules and interpolated latent embeddings (and labels) using uniformly sampling coefficient $\lambda$ from a range of [0, 1] (which is the same with the choice of $\lambda$ in our method). As shown in the table, our method indeed shows a significantly better $\Delta$ROC-AUC compared to latent mixup.

\end{document}